 \documentclass[pmlr,twocolumn,10pt]{jmlr} 

\usepackage{booktabs}
\usepackage{siunitx}

\newcommand{\equal}[1]{{\hypersetup{linkcolor=black}\thanks{#1}}}

\theorembodyfont{\upshape}
\theoremheaderfont{\scshape}
\theorempostheader{:}
\theoremsep{\newline}

 \newcommand{\bs}[1]{\boldsymbol{#1}}        

  \newcommand{\RR}[1]{\mathbb{R}^{#1}}
  \DeclareMathOperator*{\argmin}{argmin}
    \DeclareMathOperator*{\argmax}{argmax}

\usepackage{multirow}

\jmlrvolume{LEAVE UNSET} 
\jmlryear{2023}
\jmlrsubmitted{LEAVE UNSET}
\jmlrpublished{LEAVE UNSET}
\jmlrworkshop{Machine Learning for Health (ML4H) 2023} 

  \title[Generative Time Series for Complex Disease Trajectories]{Modeling 
  Complex Disease Trajectories 
  using 
  Deep 
  Generative 
  Models
  with 
  Semi-Supervised
  Latent Processes
  }

\author{%
\\
\Name{Cécile Trottet}\equal{These authors contributed equally. \\cecileclaire.trottet@uzh.ch, manuel.schuerch@uzh.ch} \\
\addr University of Zurich, ETH AI Center, Switzerland
\AND
\Name{Manuel Sch\"urch}\footnotemark[1] \\
\addr University of Zurich, ETH AI Center, Switzerland
\AND
\Name{Ahmed Allam}\\
\addr  University of Zurich, Switzerland
\AND
\Name{Imon Barua} \\
\addr  Oslo University Hospital,  University of Oslo,  Norway
\AND
\Name{Liubov Petelytska} \\
\addr University Hospital Zurich, Switzerland,  Bogomolets National Medical University, Ukraine
\AND
\Name{Oliver Distler} \\
\addr University Hospital Zurich, Switzerland
\AND
\Name{Anna-Maria Hoffmann-Vold} \\
\addr University Hospital Zurich, Switzerland, Oslo University Hospital, Norway 
\AND
\Name{Michael Krauthammer}\\
\addr University of Zurich, University Hospital Zurich, ETH AI Center, Switzerland
\AND
\Name{the EUSTAR collaborators}
}

\begin{document}

\maketitle

\begin{abstract}
In this paper, we propose a deep generative time series approach using latent temporal processes for modeling and holistically analyzing complex 
disease trajectories. 
We aim to find meaningful temporal latent representations of an underlying generative process that 
explain the observed disease trajectories 
in an interpretable and comprehensive way.
To enhance the interpretability of these latent temporal processes,
we develop a semi-supervised approach for 
disentangling
the latent space using established medical concepts.
%
By combining the generative approach with medical knowledge, we leverage the ability to 
discover 
novel aspects of the disease
while integrating medical concepts into the model.

We show that the learned temporal latent processes can be utilized for further data analysis and clinical hypothesis testing, including finding similar patients and clustering the disease into new sub-types.
Moreover, our method enables personalized online monitoring and prediction of multivariate time series including uncertainty quantification.
%
We demonstrate the effectiveness of our approach in modeling systemic sclerosis,
showcasing the potential of our machine learning model to capture 
complex disease trajectories and acquire new medical knowledge.\\ 
\end{abstract}


\section{Introduction}
\label{sec:introduction}


Understanding and analyzing 
clinical
trajectories of complex diseases - such as systemic sclerosis - is crucial for improving diagnosis, treatment, and patient outcomes. However, modeling 
such
multivariate  time series data
poses significant challenges due to the high dimensionality of clinical measurements, low signal-to-noise ratio, 
sparsity, and the complex interplay of various - potentially unobserved - factors
influencing the disease progression. 
\\ \quad \\
Therefore, 
our primary goal 
is to develop 
a machine learning (ML) model suited for the holistic analysis of
temporal disease trajectories. 
Moreover, we aim to uncover meaningful temporal latent representations capturing the complex interactions within the raw data while also providing interpretable insights, 
and potentially revealing 
novel
medical aspects of clinical disease trajectories.
%
%
%
\\ \quad \\
To achieve these goals, 
we present a deep generative temporal model that captures both the joint distribution of all the observed longitudinal clinical variables and of the latent temporal variables (\autoref{fig:overview}).
Since inferring interpretable temporal representations in a fully unsupervised way is very challenging \citep{locatello2020sober}, 
we propose a semi-supervised approach for 
disentangling
the latent space using known medical concepts
to enhance the interpretability. 
%
%
Combining an unsupervised latent generative model with known medical labels allows for the discovery of novel medically-driven patterns in the data. 
%
%
%
%
%
%
%
%
%
%
%
%
%
\\ \quad \\
Deep probabilistic generative models 
%
(Appendix \ref{sec:deep_gen} and \citeauthor{Tomczak2022DeepModeling} \citeyearpar{Tomczak2022DeepModeling})
provide a more holistic approach to modeling complex data than 
deterministic discriminative models.
%
By learning the joint distribution over all involved variables, they model the underlying generating mechanism.
In contrast, discriminative models only learn the conditional distribution of the target variable given the input variables. 
\\ \quad \\
%
%
%
While our method is general and can be applied to a wide range of high-dimensional clinical datasets, in this paper, we demonstrate its effectiveness in
modeling the progression of systemic sclerosis (SSc), a severe and yet only partially understood autoimmune disease. SSc triggers the immune system to attack the body's connective tissue, causing severe damage to the skin and multiple other internal organs. 
We seek to understand the evolution of SSc by modeling the patterns of organ involvement and progression. In doing so, we aim to learn temporal hidden representations that distinctly capture the disentangled medical concepts related to each organ.
\begin{figure}[htbp]
\floatconts
  {fig:overview}
  {\caption{Temporal generative model for systemic sclerosis.  }}
    {\includegraphics[width=.95\linewidth]{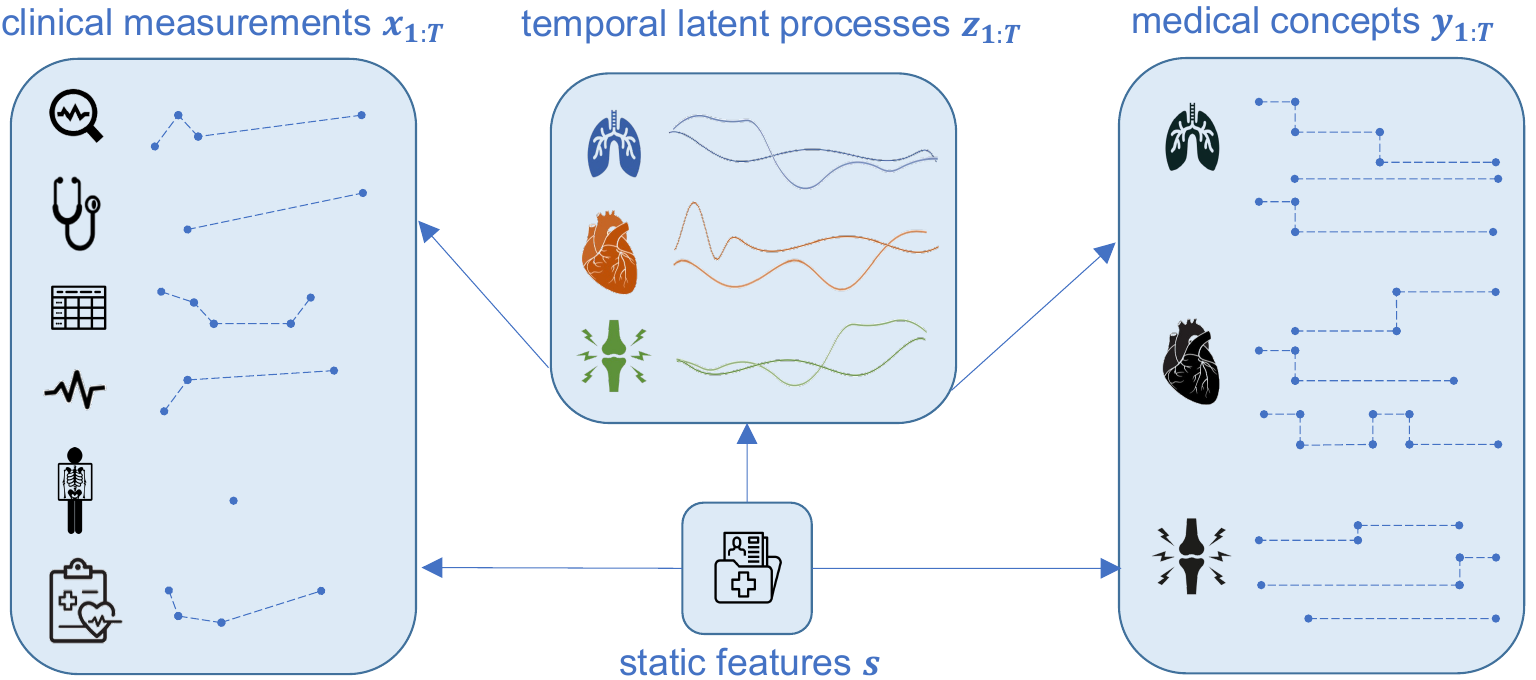}}
\end{figure}
Our approach offers several 
contributions:
\begin{itemize}
\item \textbf{Interpretable Temporal Latent Process:} 
Our generative model
allows the non-linear projection of  
patient trajectories onto a lower-dimensional temporal latent process, providing useful representations for visualization and understanding of complex medical time series data.
%

\item \textbf{Semi-Supervised 
Guided 
Latent Process:} 
To achieve more interpretable latent temporal spaces,
we propose a semi-supervised approach for 
disentangling
the latent space by leveraging known medical concepts.
  By combining the generative approach with medical domain knowledge, 
new aspects of the disease can be discovered.

\item
    \textbf{Online 
    Prediction
    with
    Uncertainty:
    }
    Our deep generative probabilistic model facilitates personalized online monitoring and robust/reliable prediction of multivariate time series data using uncertainty quantification.
%
%
    \item \textbf{Facilitating Clinical Hypothesis Testing:} The learned temporal latent processes can be inspected for further data analysis and clinical hypothesis testing, such as finding similar patients and clustering the disease trajectories into new sub-types.


      \item \textbf{Large-Scale
      Analysis of SSc
      \footnote{
      The focus of this paper is on the ML methodology, while  clinical  insights 
 about
 systemic sclerosis
 will be discussed in a clinical follow-up paper, 
 see \ref{se:clinic_ssc} for more details.
      }
     :}
      We demonstrate the potential of our ML model 
     for 
      comprehensively 
      analyzing 
      SSc for the first time at a large scale including multiple organs and various observed clinical variables.
\end{itemize}
\clearpage
\section{Background}
\label{sec:background}

\subsection{Generative Latent Variable Models}
Learning latent representations from raw 
data has a long tradition in statistics and ML with foundational research such as principal component analysis \citep{hotelling1933analysis}, factor analysis \citep{lawley1962factor} or independent component analysis \citep{comon1994independent}, which all can be used to project high-dimensional
tabular data to a latent space.  
For temporal data, models with latent processes such as hidden Markov models \citep{baum1966statistical} 
and
Gaussian processes \citep{williams2006gaussian} 
have extensively been used for discrete and continuous time applications, respectively. 
Conceptually, all these models can be interpreted as probabilistic generative models with latent variables (e.g.\ \cite{murphy2022probabilistic}), however only exploiting linear or simple mappings from the original to the latent space.

In their seminal work on Variational Autoencoders (VAEs), \citet{kingma2013auto} proposed a powerful generalization for latent generative models.
The key idea is to use deep neural networks as powerful function approximators to learn the moments of the distributions in the generative probabilistic model, enabling the representation of arbitrarily complex distributions.
Inference for the parameters of the neural networks is done with amortized variational inference (VI) \citep{blei2017variational}, a powerful approximate Bayesian inference tool.
There are various successors building and improving on the original model, for instance, conditional VAE \citep{sohn2015learning}, LVAE
\citep{sonderby2016ladder}, or VQ-VAE \citep{van2017neural}. Moreover, 
there are also several extensions that explicitly model time in the latent space such as RNN-VAE \citep{chung2015recurrent}, 
GP-VAE \citep{casale2018gaussian, fortuin2020gp}, or longitudinal VAE \citep{ramchandran2021longitudinal}.

While these approaches have showcased remarkable efficacy in generating diverse objects such as images or modeling time series, the interpretability of the resulting latent spaces or processes remains limited for complex data. Moreover, the true underlying concepts for known processes often cannot be recovered, and instead become \textit{entangled} within a single latent factor \citep{bengio2013representation}. 
Thus, there is ongoing research in designing generative models with disentangled latent factors, such as
 $\beta-$VAE \citep{higgins2016beta}, factorVAE \citep{kim2018disentangling}, TCVAE \citep{chen2018isolating} or temporal versions including
 disentangled sequential VAE \citep{hsu2017unsupervised} and
 disentangled GP-VAE \citep{bing2021disentanglement}.

However, learning interpretable and disentangled latent representations is highly difficult or even impossible for complex data without any inductive bias 
\citep{locatello2020sober}. Hence, purely unsupervised modeling falls short, leading researchers to focus on weakly supervised latent representation learning instead
\citep{locatello2020weakly, zhu2022sw, palumbo2023deep}.

In a similar spirit, we tackle the \emph{temporal} semi-supervised guidance of the latent space by providing sparse labels representing established medical domain knowledge concepts. We model the progression of complex diseases
in an unsupervised way using the raw clinical measurements, while also including medical temporal concept labels.

\subsection{Analyzing Disease Trajectories with ML}
Recently, extensive research has focused on modeling and analyzing clinical time series with machine learning --
we refer to \citet{allam2021analyzing} for a recent overview. However, most approaches focus on deterministic time series forecasting, and only a few focus on interpretable representation learning with deep models \citep{Trottet2023ExplainableDiseases} and irregularly sampled times \citep{chen2023dynamic} or on online uncertainty quantification with generative models 
\citep{schurch2020recursive, cheng2020sparse, rosnati2021mgp}.

Furthermore, prior research on data-driven analysis of systemic sclerosis is limited.
In their recent review, \citet{bonomi2022use} discuss the existing studies applying machine learning for precision medicine in systemic sclerosis. However, all of the listed studies are limited by the small cohort size (maximum of 250 patients), making the use of deep learning models challenging. Deep models were only used for analyzing imaging data (mainly nailfold capillaroscopy, \citet{garaiman2022vision}). Furthermore, most existing works solely focus on the involvement of a single organ in SSc, namely interstitial lung disease (ILD), and on forecasting methods.  To the best of our knowledge, our work is the first attempt at such a comprehensive  and large-scale (N=5673 patients)
ML analysis 
      of
      systemic sclerosis involving multiple organs and a wide range of observed clinical variables together with a systematic integration of medical knowledge.





\section{Methodology}
\label{sec:method}
We analyze patient histories that consist of two main types of data: raw temporal clinical measurements $\bs{x} = \bs{x}_{1:T}\in \RR{D \times T}$, such as 
blood pressure,
and sparse medical concept labels $\bs{y} = \bs{y}_{1:T}\in \RR{P \times T}$, describing higher-level medical definitions related to the disease, for instance, the medical definition of severity staging of the heart involvement (\autoref{fig:overview}).
The medical concept definitions are typically derived from multiple clinical measurements using logical operations. For example, a patient may be classified as having ``severe heart involvement" if certain conditions are satisfied, for instance, $\bs{x}^{(i)}>\varepsilon$ AND $\bs{x}^{(j)} = 1$. 
Both the raw measurements and labels are irregularly sampled, and we denote by $\bs{\tau}_{1:T} \in \RR{T}$ the vector of sampling time-points.
Moreover, static information denoted as $\bs{s}\in \RR{S}$ is present, alongside additional temporal covariates such as medications $\bs{p}_{1:T} \in \RR{P \times T}$  for each patient.

We condition our generative model on the context variable $\bs{c} = \{\bs{\tau}, \bs{p}, \bs{s} \}$ to be able to generate latent processes under certain conditions, for instance when a specific medication is administered. 
Furthermore, in the next sections, we introduce our approach to learning multivariate latent processes denoted as $\bs{z} = \bs{z}_{1:T}\in \RR{L \times T}$, responsible for generating both the raw clinical measurement processes $\bs{x}_{1:T}$ and the medical labels $\bs{y}_{1:T}$. In particular, we use the different temporal medical concepts to disentangle the $L$ dimensions of the latent processes by allocating distinct dimensions to represent different medical concepts.

We assume a dataset  $\{\bs{x}^i_{1:T_i}, \bs{y}^i_{1:T_i}, \bs{c}^i_{1:T_i} \}_{i=1}^N$ of $N$ patients, 
and omit the dependency to $i$  and the time index when the context is clear. 
Note that the measurements and medical concepts are often partially observed, see more details in 
Appendix \ref{sec:partially_obs}.

\subsection{Generative Model}
\label{subsec:gen_model}
We propose the probabilistic conditional generative latent variable model
\begin{align*}
    p_{\psi}(\bs{y}, \bs{x}, \bs{z} \vert \bs{c})
    =
     p_{\gamma}(\bs{y} \vert \bs{z}, \bs{c})
       p_{\pi}(\bs{x} \vert \bs{z}, \bs{c})
      p_{\phi}(\bs{z} \vert \bs{c}),
\end{align*}
with 
learnable
prior network 
$ p_{\phi}(\bs{z} \vert \bs{c})$, 
measurement likelihood network
$p_{\pi}(\bs{x} \vert \bs{z}, \bs{c})$,
and 
guidance
network 
$ p_{\gamma}(\bs{y} \vert \bs{z}, \bs{c})$,
 where 
$\psi = \{\gamma, \pi, \phi\}$ are learnable parameters.
For the sake of brevity, we do not include the time index explicitly.
Although the measurements and the concepts are conditionally independent given the latent variables, the marginal distribution 
$p_{\psi}(\bs{y}, \bs{x}\vert \bs{c}) = \int p_{\psi}(\bs{y}, \bs{x}, \bs{z} \vert \bs{c}) d \bs{z}$ allows arbitrarily rich correlations among the observed variables.
\subsection{Prior of Latent Process}
\label{sec:priorL}
We use a learnable prior network  for the latent temporal variables $\bs{z}$,
 that is,
\begin{align*}
 p_{\phi}(\bs{z} \vert \bs{c})
&
  =
 \prod_{t=1}^T 
     \prod_{l=1}^L
   \mathcal{N}\left(
   \bs{z}_t^l \vert 
\mu_{\phi}^l(\bs{c}_t), \sigma^l_{\phi}(\bs{c}_t)
   \right),
 \end{align*}
 conditioned on the context variables 
 $\bs{c}=\{\bs{\tau}, \bs{p}, \bs{s}\}$,
 so that  time-varying or demographic effects can be learned in the prior
 (Appendix 
\ref{sec:app:prior}).
%
The means $\mu_{\phi}^l(\bs{c}_t)$ and variances $\sigma^l_{\phi}(\bs{c}_t)$ are learnable deep neural networks.
We assume a factorized Gaussian prior distribution per time and latent dimensions, 
 however, many interesting extensions 
 including  continuous-time priors
 are straightforward 
 (Appendix \ref{sec:diff_prior}).
\begin{figure}[htbp]
\floatconts
  {fig:model}
  {\caption{Semi-supervised temporal latent variable model with generative and inference model.
  }}
    {\includegraphics[width=.95\linewidth]{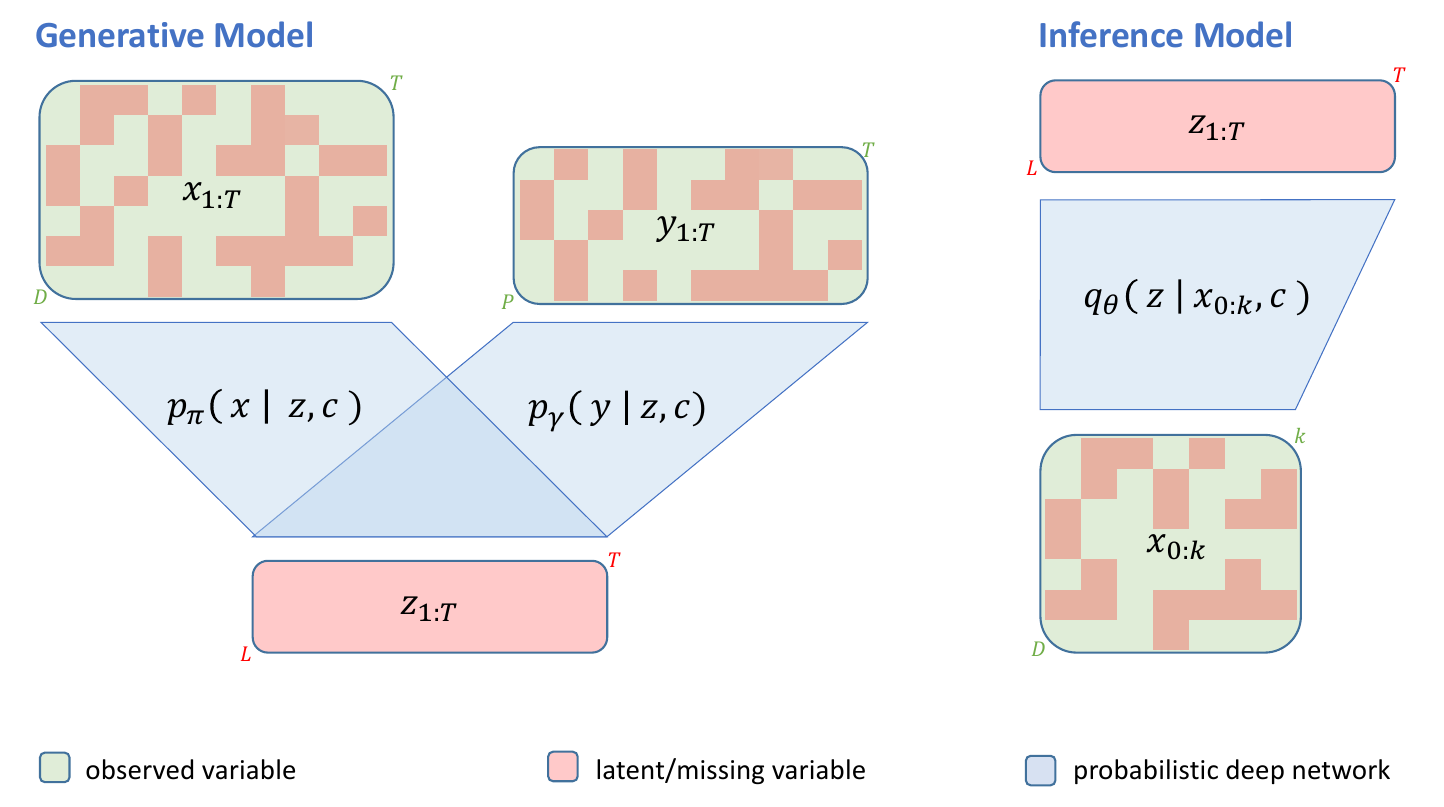}}
\end{figure}
\subsection{Likelihood of Measurements}
\label{sec:likelihood}
The probabilistic likelihood network 
maps the 
latent temporal processes $\bs{z} \in \RR{L \times T}$ together with the context variables $\bs{c}$ to the
clinical
measurements $\bs{x} \in \RR{D \times T}$, i.e.
%
\begin{align*}
 p_{\pi}(\bs{x} \vert \bs{z}, \bs{c})
 &
 =
 \prod_{t=1}^T 
  \prod_{d \in \mathcal{G}}
\mathcal{N}(x_{t}^d \vert \mu_{\pi}^d
,  \sigma^d_{\pi}
)
\prod_{d \in \mathcal{K}}
\mathcal{C}(x_{t}^d \vert p_{\pi}^d
),
\end{align*}
where we assume time- and feature-wise conditional independence.
We assume either Gaussian $\mathcal{N}$ or categorical $\mathcal{C}$ likelihoods for the observed variables $\bs{x}$, where $\mathcal{G}$ and $\mathcal{K}$ are the corresponding indices. 
%
The moments of these distributions are deep parametrized functions, i.e. for the
   mean $\mu_{\pi}^d=\mu_{\pi}^d(\bs{z}_{t}, \bs{c}_{t})$, variance    $\sigma^d_{\pi}=\sigma^d_{\pi}(\bs{z}_{t}, \bs{c}_{t})$, and
   category
  probability vector 
  $p_{\pi}^d=p_{\pi}^d(\bs{z}_{t}, \bs{c}_{t})$.
  Although the likelihood is a parametric distribution, we want to emphasize that the posterior distribution can be arbitrarily complex after marginalizing out the latent process $\bs{z}$.
%
%


\subsection{Semi-Supervised Guidance Network}
\label{sec:meth:dis}
We propose a semi-supervised approach to disentangle the latent process $\bs{z}$ with respect to defined medical concepts $\bs{y} = \bs{y}_{1:T}\in \RR{P \times T}$. 
In particular, we assume
\begin{align*}
p_{\gamma}(\bs{y} \vert \bs{z}, \bs{c})
& 
 =
 \prod_{t=1}^T 
  \prod_{g=1}^G
    \prod_{j \in \nu(g)}
   \mathcal{C}(y_{t}^j \vert h_{\gamma}^j( \bs{z}_{t}^{\varepsilon(g)}, \bs{c}_t ) ),
   %
\end{align*}
where $h_{\gamma}^j(\bs{z}_{t}^{\varepsilon(g)}, \bs{c}_t )$ is a deep parametrized probability vector, 
and $\nu(g)$ and $\varepsilon(g)$  correspond to the indices of the  $g$th  guided medical concept, and the  indices in the latent space defined for guided concept $g$,
respectively.

\subsection{Posterior of Latent Process}
\label{sec:post}
We are mainly interested in the posterior distribution $p_{\psi}(\bs{z} \vert \bs{x}, \bs{y}, \bs{c})$ of the latent process given the observations,
which we 
approximate with
an 
amortized 
variational distribution 
(Section \ref{sec:inference}, Appendix \ref{sec:app:inference})
$$ q_{\theta}(\bs{z} \vert \bs{x}, \bs{c})  \approx  p_{\psi}(\bs{z} \vert \bs{x}, \bs{y}, \bs{c}).$$
In particular, we use the 
amortized variational distribution 
\begin{align*}
q_{\theta}(\bs{z} \vert \bs{x}_{0:k}, \bs{c})
&
=
\prod_{t=1}^T
\prod_{l=1}^L  \mathcal{N}( z_{t}^l\vert \mu_{\theta}^l(\bs{x}_{0:k}, \bs{c}), \sigma^{l}_{\theta}(\bs{x}_{0:k}, \bs{c}) ) 
 \end{align*}
 with variational parameters $\theta$ and $0 \leq k \leq T$.
 Note that only the measurements $\bs{x}_{0:k}$ until observation $k$ are part of the variational distribution, and not the medical concepts $\bs{y}$. If $k=T$, there is no forecasting, whereas for $0\leq k< T$, we can also forecast the future latent variables $\bs{z}_{k+1:T}$ from the first measurements $\bs{x}_{0:k}$. 
%

\subsection{Probabilistic Inference} 
\label{sec:inference}
Since exact inference with the 
marginal likelihood
 $ p_{\psi}(\bs{x}, \bs{y} \vert \bs{c})
 = \int
  p_{\gamma}(\bs{y} \vert \bs{z}, \bs{c})
       p_{\pi}(\bs{x} \vert \bs{z}, \bs{c})
      p_{\phi}(\bs{z} \vert \bs{c})
      d\bs{z}$
 is not feasible (Appendix \ref{sec:app:inference}),
 we apply amortized variational inference \citep{blei2017variational} by maximizing a
lower bound 
$\log p_{\psi}(\bs{x}, \bs{y}\vert \bs{c}) \geq \mathcal{L}( \psi, \theta; \bs{x},  \bs{y}, \bs{c} )$
of the intractable marginal log likelihood. 
For a fixed $k$, this leads to the following objective function
\begin{align*}
    \begin{split}
 \mathcal{L}_k
  ( \psi, \theta; \bs{x},  \bs{y}, \bs{c})
  =
 & ~
 \mathbb{E}_{q_{\theta}(\bs{z} \vert \bs{x}_{0:k}, \bs{c})}\left[
 \log p_{\pi}(\bs{x} \vert \bs{z}, \bs{c}) 
  \right]
  \\
  +
  &~
\alpha ~\mathbb{ E}_{q_{\theta}(\bs{z} \vert \bs{x}_{0:k}, \bs{c})}\left[
 \log 
 p_{\gamma}(\bs{y} \vert \bs{z}, \bs{c})
  \right]
  \\
  -
  &~
 \beta ~KL\left[
q_{\theta}(\bs{z} \vert \bs{x}_{0:k}, \bs{c} )
~\vert\vert~
p_{\phi}(\bs{z} \vert \bs{c})
  \right]
  ,
   \end{split}
\end{align*}
where we introduce weights $\alpha$ and  $\beta$ inspired by the disentangled $\beta-$VAE \citep{higgins2016beta}.

The first term $\mathbb{E}_{q_{\theta}(\bs{z} \vert \bs{x}, \bs{c})}\left[
 \log p_{\pi}(\bs{x} \vert \bs{z}, \bs{c}) 
  \right]$
  is unsupervised, whereas the second
  $\alpha \mathbb{E}_{q_{\theta}(\bs{z} \vert \bs{x}, \bs{c})}\left[
 \log 
 p_{\gamma}(\bs{y} \vert \bs{z}, \bs{c})
  \right]$
  is supervised 
  and 
  $\beta KL\left[
q_{\theta}(\bs{z} \vert \bs{x}, \bs{c} )
\vert\vert
p_{\phi}(\bs{z} \vert \bs{c})
  \right]$
is a regularization term, ensuring that 
the posterior is close 
to the prior with respect to the
Kullback-Leibler (KL) divergence.
Since all dimensions in the latent space $\bs{z}$ are connected to all the measurements $\bs{x}$,
all the potential correlations between clinically measured variables can be exploited in an unsupervised fashion while disentangling the latent variables using the guidance networks for $\bs{y}$. 
The expectation over the variational distribution $\mathbb{E}_{q_{\theta}(\bs{z} \vert \bs{x}, \bs{c})}$ is approximated with a few Monte-Carlo samples (Appendix \ref{sec:app:inference}).

Given a dataset with $N$ $\mathrm{iid}$ patients $\{\bs{x}_{1:T_i}^i, \bs{y}_{1:T_i}^i, \bs{c}_{1:T_i}^i\}_{i=1}^N$,  the optimal 
parameters are given by
$$\psi^*, \theta^* = 
\argmax_{\psi, \theta}
\sum_{i=1}^N
\sum_{k=0}^{T_i}
\mathcal{L}_k( \psi, \theta; \bs{x}^i,  \bs{y}^i, \bs{c}^i ),
$$
which is computed with stochastic optimization using mini-batches of patients and different values of $k$ (Appendix \ref{sec:app:Nsamples}).
Since real-world time series data often contains many missing values, the objective function can be adapted accordingly (Appendix 
\ref{sec:partially_obs}).

\subsection{Online Prediction with Uncertainty}
\label{sec:monit}
Our model can be used for online monitoring and continuous prediction of high-dimensional medical concepts and clinical measurement distributions  
based on an increasing number of available clinical observations $\bs{x}_{0:k}$ for $k=0,1,\ldots, T$. The distributions
\begin{align*}
q_*
(\bs{y} \vert \bs{x}_{0:k}, \bs{c} ) 
&=
\int
p_{\gamma^*}(\bs{y} \vert \bs{z}, \bs{c} ) q_{\theta^*}(\bs{z} \vert \bs{x}_{0:k}, \bs{c} ) d \bs{z}
\\
q_*
(\bs{x} \vert \bs{x}_{0:k}, \bs{c} ) 
&=
\int
p_{\pi^*}(\bs{x} \vert \bs{z}, \bs{c} ) q_{\theta^*}(\bs{z} \vert \bs{x}_{0:k}, \bs{c} ) d \bs{z}
\end{align*}
are approximated 
with a two-stage Monte-Carlo sampling (Appendix \ref{sec:app:predictive_distribution}).
 
 The former can be used to automatically label and forecast the multiple medical concepts based on the raw and partially observed measurements, whereas the latter corresponds to the reconstruction and forecasting of partially observed trajectories.
 Note that these distributions represent a complex class of potentially multi-modal distributions.

\subsection{Patient Similarity and Clustering}
\label{subsec:patient_similrity}
The learned posterior network
$q_{\theta^*}(\bs{z}_{1:T} \vert \bs{x}_{1:T}, \bs{c}_{1:T} )$ can be used to map any
observed
patient 
trajectory 
$\mathcal{T}_i = \{\bs{x}_{1:T_i}^i, \bs{c}_{1:T_i}^i\}$ 
to its latent trajectory
$$\mathcal{H}_i
=
h(\mathcal{T}_i)
=
\mathbb{E}_{
q_{\theta}(\bs{z}_{1:T_i}^i \vert \bs{x}_{1:T_i}^i, \bs{c}_{1:T_i}^i)
}
\left[
 \bs{z}_{1:T_i}^i
  \right]
$$
by taking the mean of the latent process. 
These temporal latent trajectories $\{ \mathcal{H}_i \}_{i=1}^N$ of the $N$ patients in the cohort are used to define a patient similarity over the partially observed and high-dimensional original disease trajectories
$\{ \mathcal{T}_i \}_{i=1}^N$.
Through our semi-supervised generative approach, the latent trajectories effectively capture the important elements from $\bs{x}_{1:T_i}^i$ and $\bs{y}_{1:T_i}^i$, without explicitly depending on $\bs{y}_{1:T_i}^i$. Indeed, all the information related to the medical concepts is learned by $\theta$. 

Since defining a patient similarity measure between two trajectories $\mathcal{T}_i$ and $\mathcal{T}_j$ in the original space is very challenging, due to the missingness and high dimensionality, we instead propose to define it in the latent space, setting 
$$
d_{\mathcal{T}}\left(
\mathcal{T}_i, \mathcal{T}_j
\right)
=
d_{\mathcal{H}}\left(
\mathcal{H}_i, \mathcal{H}_j
\right).
$$
To measure the similarity $d_{\mathcal{H}}\left(
\mathcal{H}_i, \mathcal{H}_j
\right)$ between latent trajectories, we employ the \emph{dynamic-time-warping (dtw)} measure to account for the different lengths of the trajectories as well as the potentially misaligned disease progressions in time \citep{muller2007dynamic}. 
We then utilize the similarity measure to cluster the disease trajectories and identify similar patient trajectories
as discussed in Section
\ref{sec:clustering}.



\subsection{Modeling Systemic Sclerosis}
\label{sec:SSc}
We aim to model the overall SSc disease trajectories as well as the distinct organ involvement trajectories for patients from the European Scleroderma Trials and Research (EUSTAR) database. We provide a description of the database in Appendix \ref{sec:app:data}. 

As a proof of concept, we focus on the involvement of three important organs in SSc, namely the lung, heart, and joints (arthritis).  Each organ has two related medical concepts: \emph{involvement} and \emph{stage}. Based upon the medical definitions provided in Appendix \ref{sec:app:def}, for each of the three organs 
$\mathcal{O} : = \{lung, heart, joints\}$, we created labels signaling the organ involvement (yes/no) and severity stage ($1-4$), respectively.  We write $o(m)$, $m\in \{involvement, stage\}$, $o \in \mathcal{O}$ to refer to the corresponding medical concept for organ $o$. We project the $D=34$ and $P=11$ input features to a latent process $\bs{z}$ of dimension $L=21$. 
For each organ, we guide a distinct subset of $7$ latent processes (non-overlapping subsets), thus all of the dimensions in $\bs{z}$ are guided.
Following the notations from \ref{sec:meth:dis}, we assume
\begin{align*}
p_{\gamma}(\bs{y} \vert \bs{z}, \bs{c})
& =
 \prod_{t=1}^T 
  \prod_{\substack{o \in \mathcal{O}}}
  \prod_{\substack{m \in \\{\{inv}, \\{stage\}}\\}}
 p_{\gamma}(\bs{y}_{t}^{\nu(o(m))} \vert \bs{z}_{t}^{\varepsilon(o(m))}, \bs{c}_t).
  \\
\end{align*}
\subsection{Deep Probabilistic Networks}
\label{sec:modelchoice}

As shown in \autoref{fig:model}, our model combines several deep probabilistic networks. For the posterior $q_{\theta}(\bs{z} \vert \bs{x}_{0:k}, \bs{c})$, we implemented a temporal network with fully connected and LSTM layers \citep{hochreiter1997long}  and multilayer perceptrons for the prior $p_{\phi}(\bs{z} \vert \bs{c})$, guidance $p_{\gamma}(\bs{y} \vert \bs{z},\bs{c})$ and likelihood $p_{\pi} (\bs{x} \vert \bs{z}, \bs{c})$ networks. Implementation details are provided in Appendix \ref{sec:app:archi}.

By changing the configuration of our framework, we are able to recover well-established temporal latent variable models. For instance, if we discard the guidance networks, the model becomes similar to a deterministic RNN-AE, or probabilistic RNN-VAE if we learn the latent space distribution. Furthermore, the likelihood variance can either be learned, or kept constant as is common practice \citep{rybkin2021simple}. 
We evaluated the predictive performance of the guided model in the probabilistic and deterministic settings, with or without learning the likelihood variance.
Many further architectural choices could be explored, such as a temporal likelihood network or a GP prior, but they are beyond this paper's scope. 

\clearpage
\section{Experiments and Results}
\label{sec:experiments}

In this section, we discuss and compare different settings of our model and present experimental results 
on the EUSTAR database.

\subsection{Model Evaluation}
When referring to the model output, we use the term ``prediction" to encompass both current (reconstructed) and future values, and use the term ``forecasting" specifically when addressing future values alone.

Our model learns the optimal parameters for the likelihood $p_{\pi} (\bs{x} \vert \bs{z}, \bs{c})$ and guidance networks $p_{\gamma}(\bs{y} \vert \bs{z},\bs{c})$ and predicts the complete $\bs{x}$ and $\bs{y}$ trajectories.
As described in \autoref{sec:modelchoice}, we assessed the model's predictive performance in probabilistic and deterministic settings, and with either learning the likelihood network variance $\sigma^*$ or setting $\sigma = 1$.
\begin{figure}[htbp]
\floatconts
{fig:performance}
{\caption{Performance for $\bs{x}$ prediction. 
}}
{\includegraphics[width=\linewidth]{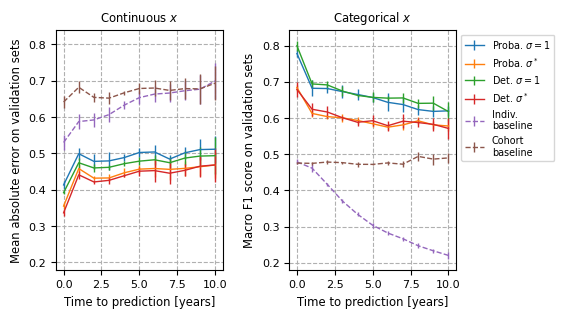}}
\end{figure}

We used $5-$fold CV 
to select the hyperparameters that achieved the lowest validation loss. Details about the inference process are provided in \autoref{sec:inference} and Appendix  \ref{sec:app:optim}. 

\autoref{fig:performance} shows the performance of predicting the clinical measurements $\bs{x}$. All of the models greatly outperform non-ML-driven individualized or cohort-based baselines. The individualized baseline predicts the patient's last available measurement for a variable as its future value. The cohort baseline predicts a value sampled from the empirical Gaussian/Categorical distribution of the variable in the cohort. The models with learned $\sigma^*$ perform slightly better at predicting continuous $\bs{x}$ while enforcing $\sigma = 1$ allows the models to better learn the categorical $\bs{x}$. The same holds for the prediction of the categorical medical concepts $\bs{y}$ (\autoref{fig:perf_y} in the appendix). Furthermore, there is no significant decrease in performance in probabilistic versus deterministic settings, even though an additional regularization term is optimized (\autoref{sec:modelchoice}). 
In Appendix \ref{sec:app:res}, we additionally compare the performance for $\bs{x}$ prediction of our guided model versus the optimal unguided baseline. 

To evaluate the uncertainty quantification, we computed the coverage of the forecasted $95\%$ confidence intervals (CI) for continuous variables and the calibration for categorical variables. Further details are provided in Appendix \ref{sec:app:res}. For continuous $\bs{x}$ forecasting, both probabilistic models achieve coverage of $92 \pm 1\%$ and of $98 \pm 0\%$ for the deterministic models, thus all slightly diverging from the optimal $95 \%$. All of the models have accurate calibration for categorical $\bs{x}$ and $\bs{y}$ forecasting (\autoref{fig:calib} in the appendix). 

The probabilistic model with learned $\sigma^*$ strikes the best balance between predictive capabilities, coverage and generative ability. In the next sections, we explore further applications and results of this final model. While the performance was computed on validation sets, the subsequent results are derived from applying the final model to a separate withheld test set. Furthermore, all of the $t-$SNE projections \citep{van2008visualizing} of the test set were obtained following the procedure described in Appendix \ref{sec:app:tsne}.
\subsubsection{Online Prediction with Uncertainty}
To illustrate how the model allows a holistic understanding of a patient's disease course, we follow an index patient $p_{idx}$ throughout the experiments. This patient has a complex disease trajectory, with varying organ involvement and stages.

We can use our model to forecast the high-dimensional distribution of $\bs{x}_{1:T}$ and $\bs{y}_{1:T}$ given the past measurements $\bs{x}_{0:k}$, as described in \autoref{sec:monit}.
For example, the heatmaps in \autoref{fig:inv_pat} show the predicted probabilities of organ involvement at a given time, overlaid with the ground truth labels. The model forecasts the probabilities after the dashed line. 
We provide online prediction plots for additional $\bs{x}$ and $\bs{y}$ in Appendix \ref{sec:app:monit}.
\begin{figure}[htbp]
\centering
\floatconts
{fig:inv_pat}
{\caption{Probabilities of organ involvement for $p_{idx}$.}}
{\includegraphics[width=.95\linewidth]{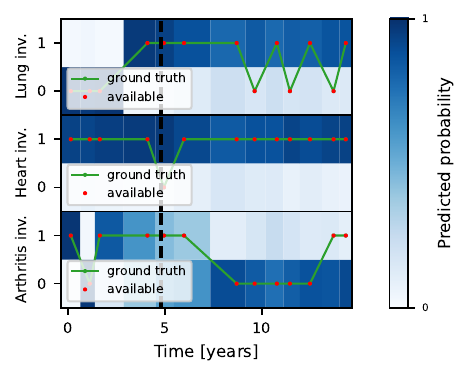}}
\end{figure}
\subsection{Cohort Analysis}
\label{sec:res_cohort}

By learning the joint distribution $p(\bs{x},\bs{y},\bs{z})$, our model allows us to analyze cohort-level disease patterns through the analysis of $\bs{z}$. Furthermore, by learning $p(\bs{z} \vert \bs{c})$, we estimate the average prior disease trajectories in the cohort. We analyze these prior trajectories in Appendix \ref{sec:app:prior}.
\subsubsection{Latent Space and Medical Concepts}


We aim to provide a method achieving semi-supervised disentanglement in the latent space. In \autoref{fig:guidance}, we compare the distribution of medical concept ground truth labels (\emph{heart stage}) in a guided versus an unguided model (i.e. without training any guidance networks). The guided model clearly provides higher medical concept disentanglement than the unguided model and thus enhances the interpretability of the different subspaces in $\bs{z}$. 
\begin{figure}[htbp]
    \floatconts
    {fig:guidance}
    {\caption{Guided versus unguided latent spaces.}}
    {\includegraphics[width=\linewidth]{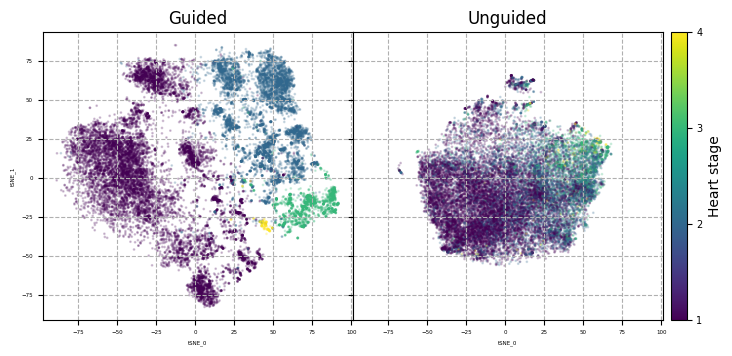}}
\end{figure}

In \autoref{fig:tsnes}, we visualize the latent space overlaid with the different predicted probabilities of organ involvement. In red, we draw the latent space trajectory of $p_{idx}$, thus getting an understandable overview of its trajectory with respect to the different medical concepts. The solid line highlights the reconstructed trajectory, whereas the dotted lines are forecasted sampled trajectories.  
\begin{figure}[htbp]
    \floatconts
    {fig:tsnes}
    {\caption{Probabilities of lung and heart involvement in the latent space.}}
    {\includegraphics[width=\linewidth]{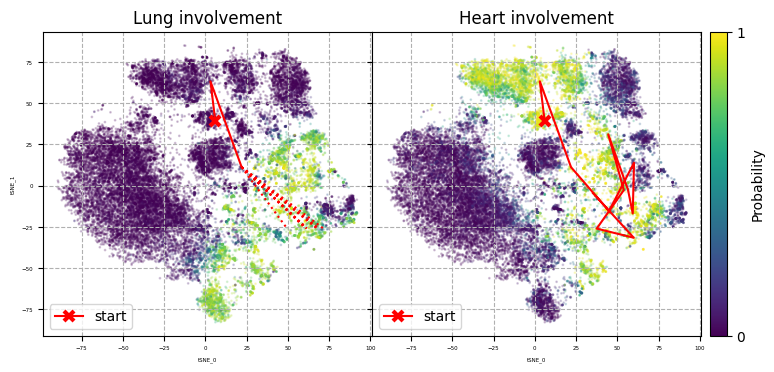}}
\end{figure}

In the first panel of \autoref{fig:tsnes}, we leverage the model's generative abilities to sample forecasted $\bs{z}$ trajectories, providing estimates of future disease phases. The model forecasts that $p_{idx}$ will move towards a region with higher probabilities of lung and heart involvement. All of the sampled trajectories converge towards the same region in this case. The second panel is overlaid with the complete reconstructed trajectory of $p_{idx}$ in $\bs{z}$. The disentanglement in the latent space enables a straightforward overview of the past and future patient trajectory. 
Additionally, \autoref{fig:tsnes_stages} in the appendix shows the patient trajectory overlaid with the predicted organ stages.

\subsubsection{Clustering and Similarity of Patient Trajectories}
\label{sec:clustering}
As described in \autoref{subsec:patient_similrity}, we compute the
dynamic-time-warping similarity measure for the latent trajectories $\mathcal{H}_i
=
h(\mathcal{T}_i)$, and subsequently apply \emph{k-means} or \emph{k-nn} to respectively cluster the multivariate time series $\{ \mathcal{H}_i \}_{i=1}^N$ or find similar patient trajectories.
We used the library implemented by
\citet{tavenard2020tslearn}. 
\begin{figure}[t]
\centering
\floatconts{fig:clust_traj_all}
  {\caption{Clustered trajectories in the latent space.}}
{\includegraphics[width=\linewidth]{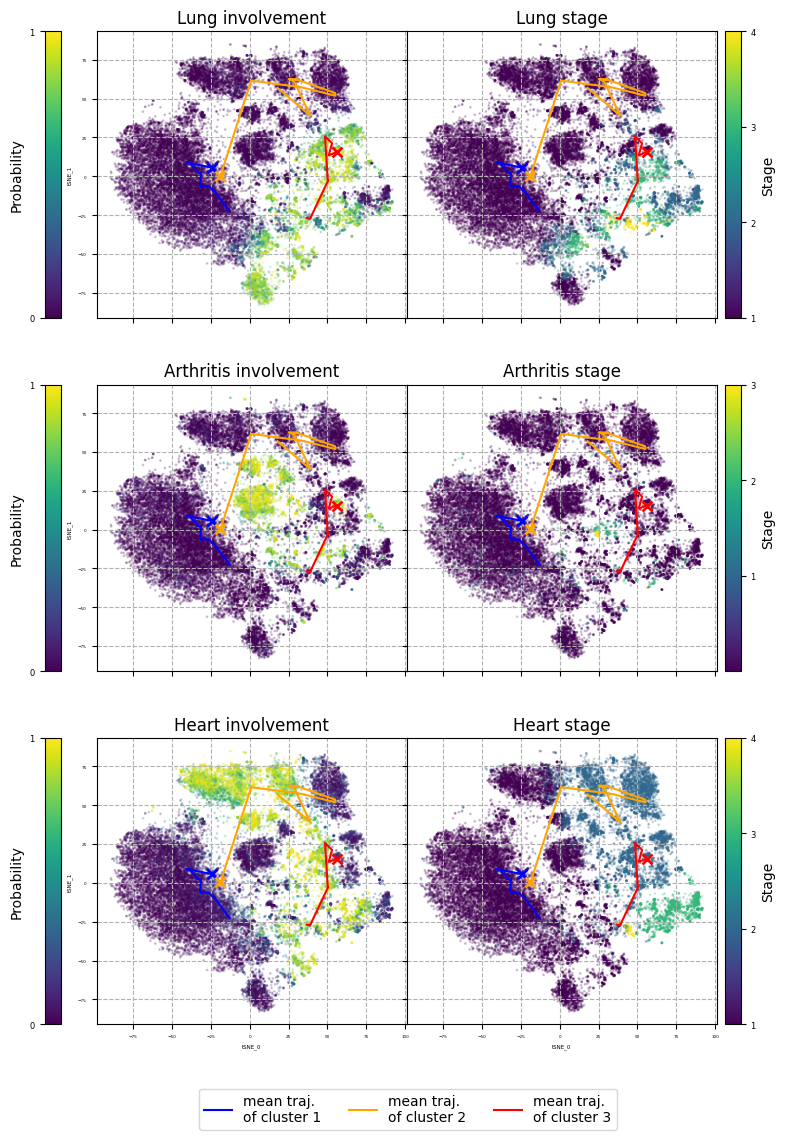}}
\end{figure}

In \autoref{fig:clust_traj_all}, we show the three mean cluster trajectories starting at the cross \textbf{x} in the latent space overlaid with the predicted medical concepts. The first found cluster corresponds to patients with no or little organ involvement. The second mean trajectory starts close to the first but progresses towards regions with heart involvement. The third cluster contains the most severely progressing patients. 
Furthermore, our modeling approach allows for direct computation of the medical concept probabilities for the mean cluster trajectories (\autoref{fig:clust_traj_proba} in the appendix). Similarly, \autoref{fig:traj_p_simil} and \autoref{fig:tsne_traj_p_simil} in the appendix compare the medical concept and latent trajectories of $p_{idx}$ and its $3$ nearest neighbors.

\section{Conclusion}
\label{sec:conclusion}

In this study, we present a novel deep semi-supervised generative latent variable approach to model complex disease trajectories. By introducing the guidance networks, we propose a  method to augment the unsupervised deep generative model with established medical concepts and achieve more interpretable and disentangled latent processes.  Our non-discriminative 
approach
effectively addresses important desiderata for healthcare models such as forecasting, uncertainty quantification, dimensionality reduction, and interpretability. Furthermore, we empirically show that our model is suited for a real-world use case, namely the modeling of systemic sclerosis, and enables a holistic understanding of the patients' disease course.
The disentangled latent space facilitates comprehensive trajectory visualizations and straightforward clustering, analysis, and forecasting of patient trajectories. 

Both our presented experiments and  modeling approaches hold the potential to be extended and adapted in many ways. We included here only the most pertinent experiments and opted for a simple architecture suited to SSc modeling. In future work, we intend to extend our framework to handle continuous time (Appendix \ref{sec:diff_prior}),
include medications for generating future hypothetical conditional trajectories (Appendix \ref{sec:app:genCondTraj}), 
include more organs in the modeling of SSc, and also include guidance networks to model additional disease dynamics like long-term outcomes. In particular, the clinical insights regarding the knowledge discovery gained from the 
clustering of trajectories such as finding new subtypes of SSc, identifying similar patients, and conducting a pathway analysis of the latent space, will be further explored in future work. 




\section{Data and Code Availability}
The dataset used is owned by a third party, the EUSTAR group, and may be obtained by request after the approval and permission from EUSTAR.
The code builds upon the pythae library \citep{chadebec2022pythae}. 
The code and examples using some artificial data are available at \url{https://github.com/uzh-dqbm-cmi/eustar_dgm4h}.

\section{Acknowledgements}
The authors thank the patients and caregivers who made the study possible, as well as all involved clinicians from the EUSTAR who collected the data. 
This work was funded by the Swiss National Science Foundation (project number 201184).

\clearpage
\bibliography{references_new}

\clearpage
\appendix
\section{Appendix}

\subsection{Clinical Insights for Systemic Sclerosis}
 \label{se:clinic_ssc}

In this paper, we present a general approach for modeling and analyzing complex disease trajectories, for which we used
 the progression of systemic sclerosis as an example.
 The focus of this paper is on the machine learning methodology, while clinically relevant
insights and data analysis regarding systemic sclerosis
will be discussed in a clinical follow-up paper where
our model will be applied to investigate the involvement of multiple organs. 

Since there is ongoing research and discussion towards finding optimal definitions of the medical concepts (involvement, stage, progression) for all impacted organs in SSc, we used preliminary definitions for three organs as a proof of concept.

\subsection{Dataset}
\label{sec:app:data}

The European Scleroderma Trials and Research group (EUSTAR) maintains a registry dataset of about 20’000 patients extensively documenting organ involvement in SSc. It contains around $30$ demographic variables, and 500 temporal clinical measurement variables documenting the patients' overall and organ-specific disease evolution. For a detailed description of the database, we refer the reader to \citet{meier2012update, hoffmann2021progressive}.

For our analysis, we included $5673$ patients with enough temporality (i.e. at least $5$ medical visits). We used $10$ static variables related to the patients' demographics and $34$ clinical measurement variables, mainly related to the monitoring of the lung, heart, and joints in SSc.  In future work, we plan to include more patients and more clinical measurements for analyzing all involved organs.
\subsection{Medical Concepts Definitions}
\label{sec:app:def}
Defining the organ involvement and stages in SSc is a challenging task as varying and sometimes contradicting definitions are used in different studies. However, there is ongoing research to find the most accurate definitions. Since this work is meant as a proof of concept, we used the following preliminary definitions of involvement and stage for the lung, heart, and joints (arthritis). The medical concepts are defined for the variables of the EUSTAR database. There are $4$ stages of increasing severity for each organ. If multiple definitions are satisfied, the most severe stage is selected. Furthermore, there is missingness in the labels due to incomplete clinical measurements. Our modeling approach thus also could be used to label the medical concepts when missing. 

We use the following abbreviations: 
\begin{itemize}
    \item Interstitial Lung Disease: ILD
    \item High-resolution computed tomography: HRCT
    \item Forced Vital Capacity: FVC
    \item Left Ventricular Ejection Fraction: LVEF
    \item Brain Natriuretic Peptide: BNP
    \item N-terminal pro b-type natriuretic peptide: NTproBNP
    \item Disease Activity Score 28: DAS28 
\end{itemize}
\subsubsection{Lung}
\paragraph{Involvement}
At least one of the following must be present: 
\begin{itemize}
    \item ILD on HRCT
    \item FVC $< 70 \%$
\end{itemize}
\paragraph{Severity staging}

\begin{enumerate}
    \item FVC $>80 \%$ or Dyspnea stage of $2$
    \item ILD extent $<20 \%$ or $70 \% < \text{FVC} \leq 80 \%$ or Dyspnea stage of $3$
    \item ILD extent $>20 \%$ or $50\% \leq \text{FVC} \leq 70 \%$ or Dyspnea stage of $4$
    \item FVC$<50\%$ or Lung transplant or Dyspnea stage of $4$
\end{enumerate}

\subsubsection{Heart}
\paragraph{Involvement}
At least one of the following must be present:
\begin{itemize}
    \item LVEF $<45\%$
    \item Worsening of cardiopulmonary manifestations within the last month
    \item Abnormal diastolic function
    \item Ventricular arrhythmias
    \item Pericardial effusion on echocardiography
    \item Conduction blocks
    \item BNP $>35$ pg/mL
    \item NTproBNP$>125$ pg/mL

\end{itemize}
\paragraph{Severity staging}
\begin{enumerate}
    \item Dyspnea stage of $1$
    \item Dyspnea stage of $2$
    \item Dyspnea stage of $3$
    \item Dyspnea stage of $4$
\end{enumerate}
\subsubsection{Arthritis}
\paragraph{Involvement}
At least one of the following must be present:
\begin{itemize}
    \item Joint synovitis
    \item Tendon friction rubs
\end{itemize}
\paragraph{Severity staging}
\begin{enumerate}
    \item DAS28 $<2.7$
    \item $2.7 \leq \text{DAS28} \leq 3.2$
    \item $3.2 < \text{DAS28} \leq 5.1$
    \item DAS28 $>5.1$
\end{enumerate}

\section{Details and Extensions for Generative Model}
\label{subsec:gen_model_extension}
In this section, we provide  more details and several possible extensions to the main temporal generative model presented in Section
\ref{subsec:gen_model}.

\subsection{Inference}
\label{sec:app:inference}
In this section, we explain the inference process of the proposed generative model $p_{\psi}(\bs{y}, \bs{x}, \bs{z} \vert \bs{c})=
p_{\gamma}(\bs{y} \vert \bs{z}, \bs{c})
       p_{\pi}(\bs{x} \vert \bs{z}, \bs{c})
      p_{\phi}(\bs{z} \vert \bs{c})$
in more detail.
We are particularly interested in the posterior of the latent variables $\bs{z}$
given  $\bs{y}$, $\bs{x}$, and $\bs{c}$, that is,
\begin{align*}
    p_{\psi}(\bs{z} \vert \bs{y}, \bs{x},  \bs{c})
    =
    \frac{p_{\psi}(\bs{y}, \bs{x}, \bs{z} \vert \bs{c})}
    {p_{\psi}(\bs{y}, \bs{x}\vert \bs{c})}
    =
    \frac{p_{\psi}(\bs{y}, \bs{x}, \bs{z} \vert \bs{c})}
    {\int p_{\psi}(\bs{y}, \bs{x}, \bs{z} \vert \bs{c}) d \bs{z} },
\end{align*}
which is in general intractable due to the marginalization of the latent process in the marginal likelihood 
$p_{\psi}(\bs{y}, \bs{x}\vert \bs{c}) = \int p_{\psi}(\bs{y}, \bs{x}, \bs{z} \vert \bs{c}) d \bs{z}$.
Therefore, we resort to approximate
inference, in particular, amortized variational inference (VI) \citep{blei2017variational}, where a variational
distribution $q_{\theta}(\bs{z} \vert \bs{x}, \bs{c})$ close to the true posterior distribution
$p_{\psi}(\bs{z} \vert \bs{x}, \bs{y}, \bs{c}) \approx q_{\theta}(\bs{z} \vert \bs{x}, \bs{c})$ is introduced. The similarity between these distributions is usually measured
in terms of KL divergence \citep{murphy2022probabilistic}, therefore, we aim to find parameters 
satisfying
$$
\theta^*, \psi^*
=
\argmin_{\theta, \psi} 
KL\left[
q_{\theta}(\bs{z} \vert \bs{x}, \bs{c})
~\vert\vert~
p_{\psi}(\bs{z} \vert \bs{x}, \bs{y}, \bs{c})
\right].
$$
This optimization problem is equivalent  \citep{murphy2022probabilistic} to maximizing a lower bound 
$\mathcal{L}( \psi, \theta; \bs{x},  \bs{y}, \bs{c} ) \leq
p_{\psi}(\bs{y}, \bs{x}\vert \bs{c})
$ 
to the intractable marginal likelihood, that is,
$$
\theta^*, \psi^*
=
\argmax_{\theta, \psi} 
\mathcal{L}( \psi, \theta; \bs{x},  \bs{y}, \bs{c} ).
$$
In particular, this lower bound equals
\begin{align*}
 \mathcal{L}
     =
    &\int
    q_{\theta}(\bs{z} \vert \bs{x}, \bs{c})
    \log
    \frac{p_{\psi}(\bs{y}, \bs{x}, \bs{z} \vert \bs{c})}
    {q_{\theta}(\bs{z} \vert \bs{x}, \bs{c})}
    d \bs{z}
    \\
    = 
    &
    \int
    q_{\theta}(\bs{z} \vert \bs{x}, \bs{c})
    \log
    \frac{p_{\gamma}(\bs{y} \vert \bs{z}, \bs{c})
       p_{\pi}(\bs{x} \vert \bs{z}, \bs{c})
      p_{\phi}(\bs{z} \vert \bs{c})}
    {q_{\theta}(\bs{z} \vert \bs{x}, \bs{c})}
    d \bs{z},
\end{align*}
which can be rearranged to
\begin{align*}
    \mathcal{L}
     =
     &~
     \mathbb{E}_{q_{\theta}(\bs{z} \vert \bs{x}, \bs{c})}\left[
 \log p_{\pi}(\bs{x} \vert \bs{z}, \bs{c}) 
  \right]
  \\
  +
  &~
 \mathbb{ E}_{q_{\theta}(\bs{z} \vert \bs{x}, \bs{c})}\left[
 \log 
 p_{\gamma}(\bs{y} \vert \bs{z}, \bs{c})
  \right]
  \\
  -
  &~
 KL\left[
q_{\theta}(\bs{z} \vert \bs{x}, \bs{c} )
~\vert\vert~
p_{\phi}(\bs{z} \vert \bs{c})
  \right].
\end{align*}
For the Gaussian prior and approximate posterior described in
Section 
\ref{sec:priorL} and \ref{sec:post}, respectively, the KL-term can be computed analytically and efficiently
\citep{Tomczak2022DeepModeling}.
On the other hand, the expectations $ \mathbb{ E}_{q_{\theta}}$
can be approximated with a few Monte-Carlo samples
$\bs{z}^1,\ldots,\bs{z}^s,\ldots,\bs{z}^S
\sim
q_{\theta}(\bs{z} \vert \bs{x}, \bs{c})
$ leading to
\begin{align*}
 \mathbb{E}_{q_{\theta}(\bs{z} \vert \bs{x}, \bs{c})}\left[
 \log p_{\pi}(\bs{x} \vert \bs{z}, \bs{c}) 
 p_{\gamma}(\bs{y} \vert \bs{z}, \bs{c})
  \right]
  \\
  \approx
  \frac{1}{S}
  \sum_{s=1}^S
   \log p_{\pi}(\bs{x} \vert \bs{z}^s, \bs{c}) 
 p_{\gamma}(\bs{y} \vert \bs{z}^s, \bs{c}).
\end{align*}

\subsubsection{
Partially Observed Data }
\label{sec:partially_obs}
The measurements 
$\bs{x}\in \RR{D \times T}$ and the concepts  $\bs{y}\in \RR{P \times T}$ contain  many missing values. We define the  indices
$\bs{o}_x \in \RR{D \times T}$ and $\bs{o}_y \in \RR{P \times T}$ for which the observations are actually measured. Therefore, 
we compute the lower bound only on the observed variables, i.e.\
$\log p_{\psi}(\bs{x}^{\bs{o}_x}, \bs{y}^{\bs{o}_y} \vert \bs{c}) \geq \mathcal{L}( \psi, \theta; \bs{x}^{\bs{o}_x},  \bs{y}^{\bs{o}_y}, \bs{c} ) $,
as is similarly done by \citet{fortuin2020gp, ramchandran2021longitudinal}.
This then leads for instance to
\begin{align*} 
\mathbb{E}_{q_{\theta}(\bs{z} \vert \bs{x}, \bs{c})}\left[
 \log p_{\pi}(\bs{x}^{\bs{o}_x} \vert \bs{z}, \bs{c}) p_{\gamma}(\bs{y}^{\bs{o}_y} \vert \bs{z}, \bs{c})
 \right],
 \end{align*}
where the related
log-likelihood
$\log p_{\pi}(\bs{x}^{\bs{o}_x} \vert \bs{z}, \bs{c})
 =
 \log \prod_{t,d \in \bs{o}_x}
 p_{\pi}(x_{t}^d \vert \bs{z}_{t}, \bs{c}_t)
 =
  \sum_{t,d \in \bs{o}_x}
 \log p_{\pi}(x_{t}^d \vert \bs{z}_{t}, \bs{c}_t)
 $
is only summed over the actually observed measurements. The same can be derived for the medical concepts $\bs{y}^{\bs{o}_y} $.

\subsubsection{
Lower Bound for N Samples }
\label{sec:app:Nsamples}
Given a dataset with $N$ $\mathrm{iid}$ patients $\mathcal{D} = \{\mathcal{D}_i\}_{i=1}^N= \{\bs{x}_{1:T_i}^i, \bs{y}_{1:T_i}^i, \bs{c}_{1:T_i}^i\}_{i=1}^N$, 
the lower bound to the marginal log-likelihood is
$$
\log p_{\psi}(\mathcal{D}) 
=
\log \prod_{i=1}^N p_{\psi}(\mathcal{D}_i) 
\geq
\sum_{i=1}^N
\mathcal{L}( \psi, \theta; \bs{x}^i,  \bs{y}^i, \bs{c}^i ),
$$
which is maximized through stochastic optimization with mini-batches (\autoref{sec:inference}).
Moreover, 
suppose we have $T+1$ iid copies of the whole dataset $\{ \mathcal{D}^{k}\}_{k=0}^T$, then
\begin{align*}
\log p_{\psi}(\{ \mathcal{D}^{k}\}_{k=0}^T) 
=
\log \prod_{i=1}^N  \prod_{k=0}^T p_{\psi}(\mathcal{D}_i^k) 
\\
\geq
\sum_{i=1}^N
\sum_{k=0}^T
\mathcal{L}_k( \psi, \theta; \bs{x}^{i,k},  \bs{y}^{i,k}, \bs{c}^{i,k} ),
\end{align*}
where 
$\mathcal{L}_k( \psi, \theta; \bs{x}^{i,k},  \bs{y}^{i,k}, \bs{c}^{i,k} )$
is the lower bound obtained by plugging in the corresponding
approximate posterior
$q_{\theta}(\bs{z} \vert \bs{x}_{0:k}, \bs{c})$.

\subsubsection{Predictive Distributions}
\label{sec:app:predictive_distribution}

The predictive distributions for the measurement $\bs{x}_{1:T}$ and concept trajectories $\bs{y}_{1:T}$ in
\autoref{sec:monit} can be obtained via a two-stage Monte-Carlo approach. For instance, we can sample from the distribution of the measurements 
\begin{align*}
&q_*
(\bs{x}_{1:T} \vert \bs{x}_{0:k}, \bs{c} ) 
\\
=
&\int
p_{\pi^*}(\bs{x}_{1:T} \vert \bs{z}_{1:T}, \bs{c} ) q_{\theta^*}(\bs{z}_{1:T} \vert \bs{x}_{0:k}, \bs{c} ) d \bs{z}
\end{align*}
by first sampling from the latent trajectories
$$
\bs{z}_{1:T}^1,\ldots,\bs{z}_{1:T}^s,\ldots\bs{z}_{1:T}^S
\sim
q_{\theta^*}(\bs{z}_{1:T} \vert \bs{x}_{0:k}, \bs{c} ) $$
given the current observed measurements
$ \bs{x}_{1:k}$.
In a second step, for each of the samples, we compute 
$$
\bs{x}_{1:T}^1,\ldots,\bs{x}_{1:T}^u,\ldots\bs{x}_{1:T}^U
\sim
p_{\pi^*}(\bs{x}_{1:T} \vert \bs{z}_{1:T}^s, \bs{c} )$$ to represent
the overall uncertainty of the measurement distribution. 

\subsection{Different Prior}
\label{sec:diff_prior}

The factorized prior described in \autoref{sec:priorL} can be extended to continuous time with Gaussian processes  (GPs, \cite{williams2006gaussian, schurch2020recursive, schurch2023correlated}), as introduced 
 by
 \cite{casale2018gaussian, fortuin2020gp} in the unsupervised setting.
 In particular, 
 we can replace
 \begin{align*}
 p_{\phi}(\bs{z} \vert \bs{c})
&
=
 p_{\phi}(\bs{z}_{1:T} \vert \bs{c}_{1:T})
 =
 \prod_{t=1}^T 
  \prod_{l=1}^L
  p_{\phi}(\bs{z}_t^l \vert \bs{c}_t)
  \\
  &
  =
 \prod_{t=1}^T 
     \prod_{l=1}^L
   \mathcal{N}\left(
   \bs{z}_t^l \vert 
\mu_{\phi}^l(\bs{c}_t), \sigma^l_{\phi}(\bs{c}_t)
   \right),
 \end{align*}
 with
 \begin{align*}
 p_{\phi}(\bs{z}_{1:T} \vert \bs{c}_{1:T})
  =
     \prod_{l=1}^L
   \mathcal{GP}\left(
   \bs{z}^l \vert 
m_{\phi}^l(\bs{c}), k^l_{\phi}(\bs{c}, \bs{c}')
   \right)
 \end{align*}
 with a mean function $m_{\phi}^l(\bs{c})$ and kernel $k^l_{\phi}(\bs{c}, \bs{c}')$, to take into account all the probabilistic correlations occurring in continuous time. This leads to a \textit{stochastic} dynamic process, which theoretically matches the assumed disease process more adequately than a deterministic one. A further advantage is the incorporation of prior knowledge via the choice of the particular kernels for each latent process so that different characteristics such as long and small lengthscales, trends, or periodicity can be explicitly enforced in the latent space. 


\subsection{Conditional Generative Trajectory Generation}
\label{sec:app:genCondTraj}
Our generative approach is also promising for conditional generative trajectory sampling, in a similar spirit as proposed by \cite{schurch2023generating}. In particular, if we use medications as additional covariates 
$\bs{p}=\bs{p}_{1:T}=\{ \bs{p}_{0:k}, \bs{p}_{k+1:T}\}$ in our approximate posterior distribution
$q_{\theta}(\bs{z} \vert \bs{x}_{0:k}, \bs{c})=
q_{\theta}(\bs{z} \vert \bs{x}, \bs{\tau}, \bs{s}, \bs{p}_{0:k}, \bs{p}_{k+1:T})$
with $\bs{c} = \{\bs{\tau}, \bs{s}, \bs{p}\}$,
 the model can be used to sample future hypothetical trajectories 
 $\bs{x}_{k+1:T}$
 with 
%
 \begin{align*}
q_*
(\bs{x}_{k+1:T}  \vert \bs{x}_{0:k}, \bs{\tau}, \bs{s}, \bs{p}_{0:k}, \bs{p}_{k+1:T} ) 
\\
=
\int
p_{\pi^*}(\bs{x}_{k+1:T} \vert \bs{z}, \bs{\tau}, \bs{s}, \bs{p}_{0:k}, \bs{p}_{k+1:T} )
\\
q_{\theta^*}(\bs{z} \vert \bs{x}_{0:k}, \bs{\tau}, \bs{s}, \bs{p}_{0:k}, \bs{p}_{k+1:T}) 
d \bs{z}
\end{align*}
 based
 on
 future query
 medications
  $\bs{p}_{k+1:T}$.

\section{Model Implementation}

\subsection{Model Architecture}
\label{sec:app:archi}
We describe the architecture and inputs/outputs of the different neural networks in our final model for SSc. For a patient with measurement time points $\bs{\tau}_{1:T}$ of the complete trajectory, the model input at time $t \in \bs{\tau}$ are the static variables $\bs{s}$, the clinical measurements $\bs{x}_{0:t}$, and the trajectory time points $\bs{\tau}$. Thus for SSc modeling, we have that $\bs{c}=\{ \bs{\tau}, \bs{s} \}$. The model $\mathcal{M}$ outputs the distribution parameters of the clinical measurements and the organ labels for all trajectory time points $\bs{\tau}$. Without loss of generality, we assume that $\bs{x}^{1:M}$ are continuous variables and $\bs{x}^{M+1:D}$ categorical, so that the model can be described as
\begin{align*}
        \mathcal{M}: \left(\bs{c},\bs{x}_{0:t}\right) \longrightarrow \\ \left(\hat{\bs{\mu}}_{1:T}^{x^{1:M}}(t), \hat{\bs{\sigma}}_{1:T}^{x^{1:M}}(t), \hat{\bs{\pi}}_{1:T}^{x^{M+1:D}}(t), \hat{\bs{\pi}}_{1:T}^{y}(t) \right).
\end{align*}
We explicitly include the dependencies to $t$ to emphasize that the parameters of the whole trajectory are estimated given the information up to time $t$.
\begin{itemize}
    \item \textbf{Prior network}: The prior is a multilayer perceptron (MLP). It takes as input $\bs{c}$ and outputs the estimated mean and variance of the prior latent distribution $\hat{\mu}_{1:T}^{prior}$ and $\hat{\sigma}_{1:T}^{prior}$.  
    \item \textbf{Encoder network} (posterior): The encoder contains LSTM layers followed by fully connected feed-forward layers. It takes as input $\bs{x}_{0:t}$ and $\bs{c}$ and outputs the estimated mean and standard deviation of the posterior distribution of the latent variables $\hat{\mu}_{1:T}^{post}(t)$ and $\hat{\sigma}_{1:T}^{post}(t)$, from which we sample the latent variables $\bs{z}_{1:T}(t)$ (complete temporal latent process) given the information up to $t$.  
    \item \textbf{Decoder network} (likelihood): The decoder is an MLP and takes as input the sampled latent variables $\bs{z}_{1:T}(t)$ and $\bs{c}$ and outputs the estimated means and standard deviations $\hat{\mu}_{1:T}^{x^{1:M}}(t)$ and $\hat{\sigma}_{1:T}^{x^{1:M}}(t)$ of the distribution of the continuous clinical measurements and class probabilities $\hat{\pi}_{1:T}^{x^{M+1:D}}(t)$ of the categorical measurements.
    \item \textbf{Guidance networks}: For each organ, we define one MLP guidance network per related medical concept (involvement and stage). A guidance network for organ 
    $o \in \mathcal{O}:= \{lung, heart, joints \}$ and related medical concept 
    $m \in \{inv, stage \}$, takes as input the sampled latent variables $\bs{z}_{1:T}^{\epsilon(o(m))}(t)$ and outputs the predicted class probabilities  $\hat{\pi}_{1:T}^{y^{\nu(o(m))}}(t)$ of the labels, where $\nu(o(m))$ are the indices in $y$ related to the medical concept $o(m)$, and $\epsilon(o(m))$ the indices in the latent space. 
\end{itemize} 


\subsection{Training Objective}

We follow the notation introduced in Section \ref{sec:method} and Appendix \ref{subsec:gen_model_extension}. To train the model to perform forecasting, for each patient, we augment the data by assuming $T+1$ \emph{iid} copies of the data $x$ and $y$ (see also \ref{sec:app:Nsamples}) and recursively try to predict the last $T - t$, $t=0,...,T$ clinical measurements and medical concepts.
The total loss for a patient $p$ is
\begin{equation}
\mathcal{L}_p = \sum_{t=0}^{T} \mathcal{L}(t), 
\end{equation}
where 
\begin{align*}
       \mathcal{L}(t) := NLL \left(\hat{\mu} ^{x^{1:M}} (t), \hat{\sigma}^{x^{1:M}} (t), \bs{x}^{1:M} \right) \\
       + CE \left( \hat{\pi} ^ {x^{M+1:D}}(t), \bs{x}^{M+1:D} \right) \\ + \alpha * CE \left( \hat{\pi}^y(t), \bs{y} \right) \\ + \beta * KL \left( \hat{\mu}^{prior}, \hat{\sigma} ^{prior}, \hat{\mu}^{post}(t), \hat{\sigma}^{post}(t)\right),
\end{align*}
where $NLL$, $CE$ and $KL$ are the negative log-likelihood, cross-entropy and KL divergence, respectively. Further, $\alpha$ and $\beta$ are hyperparameters weighting the guidance and KL terms.

\subsubsection{Model Optimization}
\label{sec:app:optim}
We only computed the loss with respect to the available measurements. We randomly split the set of patients $\mathcal{P}$ into a train set $\mathcal{P}_{train}$ and test set $\mathcal{P}_{test}$  and performed $5-$fold CV with random search on $\mathcal{P}_{train}$ for hyperparameter tuning. Following the principle of empirical risk minimization, we trained our model to minimize the objective loss over $\mathcal{P}_{train}$, using the Adam \citep{kingma2014adam} optimizer with mini-batch processing and early stopping. 

\subsubsection{Architecture and Hyperparameters}
We tuned the dropout rate and the number and size of hidden layers using $5$-fold CV, and used a simple architecture for our final model. The posterior network contains a single lstm layer with hidden state of size $100$, followed by two fully connected layers of size $100$. The likelihood network contains two separate fully connected layers of size $100$, learning the mean and variances of the distributions separately. The guidance networks contain a single fully connected layer of size $40$ and the prior network a single fully connected layer of size $50$. We used batch normalization, ReLU activations, and a dropout rate of $0.1$. We set $\alpha = 0.2$ and $\beta = 0.01$. 
\section{Results}
\subsection{Model Evaluation}
\label{sec:app:res}
We discuss the evaluation results for unguided models, medical concept prediction, and uncertainty quantification. In \autoref{fig:perf_x_bsl}, we compare the performance of the clinical measurement $\bs{x}$ prediction of the different guided models versus their unguided counterparts (with the same number of latent processes). Note that these unguided models are optimal baselines for $\bs{x}$ prediction since they are not trained to predict $\bs{y}$, too. As \autoref{fig:perf_x_bsl} shows, the unguided models usually outperform the guided models, but the difference is not significant for the probabilistic models. Unsurprisingly, the best performing model is a deterministic unguided model, i.e. not trained to learn the $\bs{z}$ and $\bs{y}$ distributions.
\begin{figure}[htbp]

\floatconts
  {fig:perf_x_bsl}
  {\caption{Performance for $x$ prediction, guided versus unguided models.}}
  {\includegraphics[width=\linewidth]{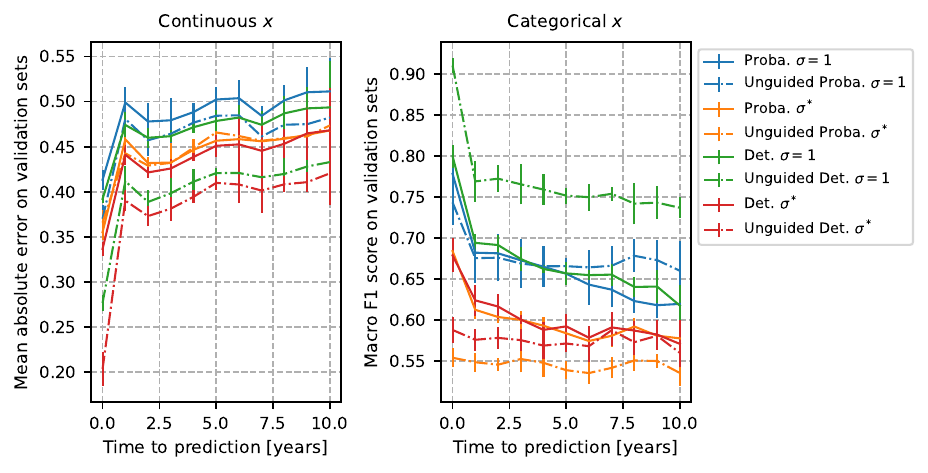}}
\end{figure}

\autoref{fig:perf_y} shows the macro $F_1$ scores for the medical concepts $\bs{y}$ prediction of the different models. The models with fixed likelihood variance generally slightly outperform the models with learned variance. All of the models outperform the individualized and cohort baselines. 
\begin{figure}[t]
\floatconts
  {fig:perf_y}
  {\caption{Performance for $y$ prediction.}}
  {\includegraphics[width=0.8\linewidth]{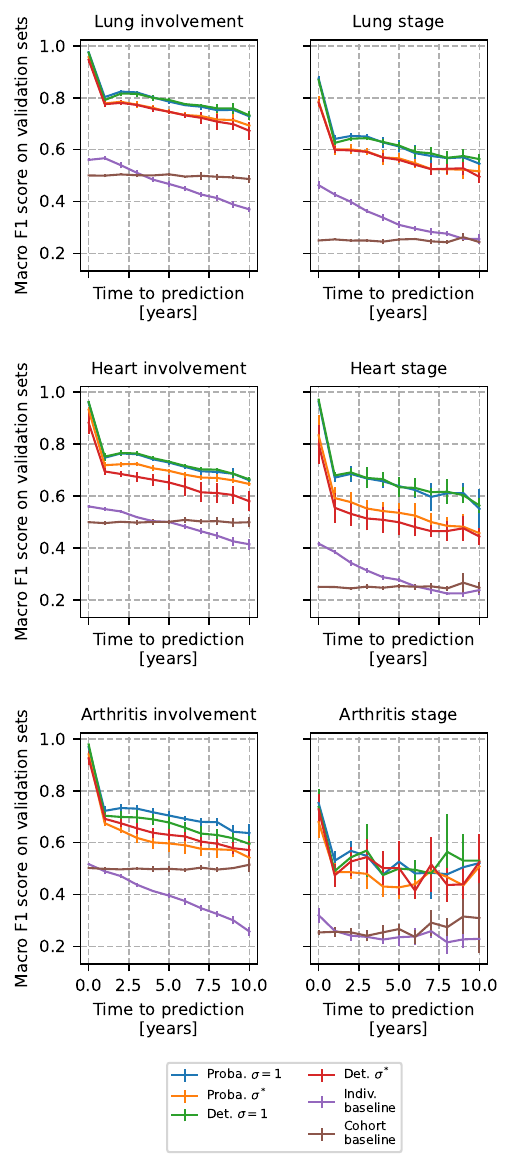}}
\end{figure}

To evaluate the uncertainty quantification of the models, we computed the coverage of the continuous predictions and calibration of the predicted probabilities for categorical measurements. The coverage is the probability that the confidence interval (CI) predicted by the model contains the true data point. Since the likelihood distribution is Gaussian, the 95\% CI is $\mu_{pred} \pm 1.96 \sigma_{pred}$. To achieve perfect coverage of the $95 \%$ CI, the predictions should fall within the predicted CI $95\%$ of the time. We computed the coverage over all forecasted data points. 
For categorical measurements, the calibration curve is computed to assess the reliability of the predicted class probabilities. They are computed in the following way. We grouped all of the forecasted probabilities (for one-hot encoded vectors) into $n=20$ bins dividing the 0-1 interval. Then, for each bin, we compared the observed frequency of ground truth positives (aka ``fraction of positive") with the average predicted probability within the bin. Ideally, these two quantities should be as close as possible, i.e. close to the line of ``perfect calibration" in \autoref{fig:calib}.
The calibration curves in \autoref{fig:calib} show that all of the models are well calibrated both in their categorical $\bs{x}$ and medical concept $\bs{y}$ forecasts (averaged over all forecasted data points in the respective validation sets). 
\begin{figure}[t]
\floatconts
{fig:calib}
{\caption{Calibration curves.}}
{
\subfigure[Categorical clinical measurements $x$][c]{\label{fig:calib_x}
\includegraphics[width = 0.8\linewidth]{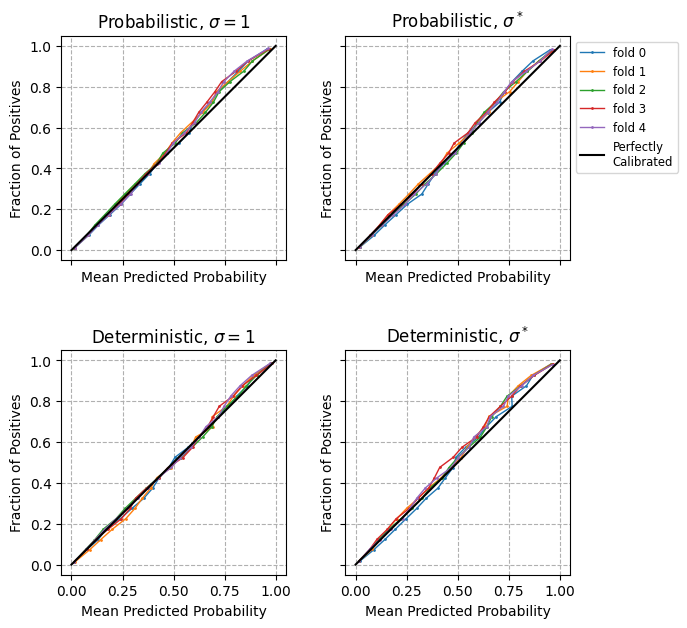}}

\subfigure[Medical concepts $y$][c]{\label{fig:calib_y} \includegraphics[width = 0.8\linewidth]{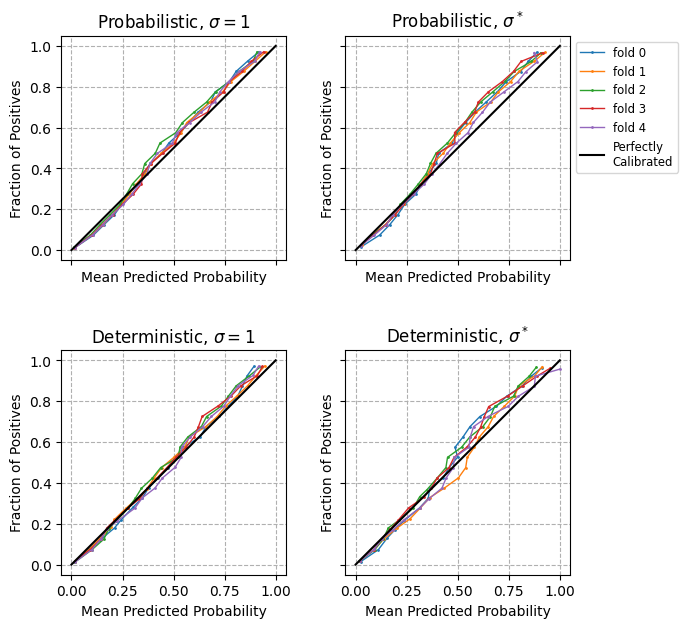}}
}
\end{figure}
\subsubsection{Online Prediction with Uncertainty}
\label{sec:app:monit}
We provide additional online prediction results for the index patient $p_{idx}$.

Figures \ref{fig:fvc} and \ref{fig:dlco_pat} show the evolution in the predicted mean and $95\%$  CI of the Forced Vital Capacity (FVC)\footnote{FVC is the amount of air that can be exhaled from the lungs.} and DLCO(SB)\footnote{DLCO(SB) stands for single breath (SB) diffusing capacity of carbon monoxide (DLCO).} for $p_{idx}$. 

The values after the dashed line are forecasted. As more prior information becomes available to the model, the forecast becomes more accurate and the CI shrinks. 
\begin{figure}[htbp]
\centering
\floatconts
{fig:fvc}
{\caption{FVC of $p_{idx}$: predicted mean and $95\%$ CI}}
{\includegraphics[width=0.6\linewidth]{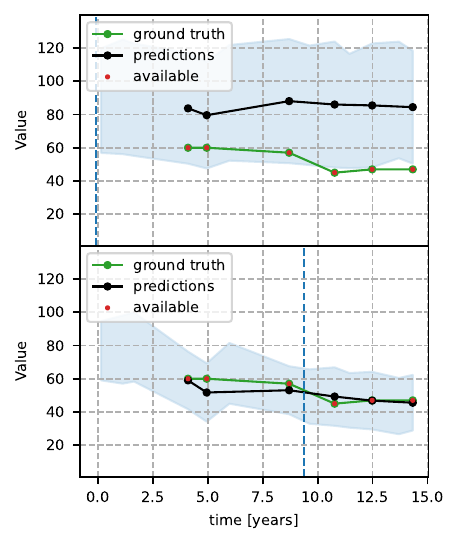}}
\end{figure}
\begin{figure}[htbp]

\floatconts
  {fig:dlco_pat}
  {\caption{DLCO(SB) of $p_{idx}$: predicted mean and $95\%$ CI}}
  {\includegraphics[width=0.6\linewidth]{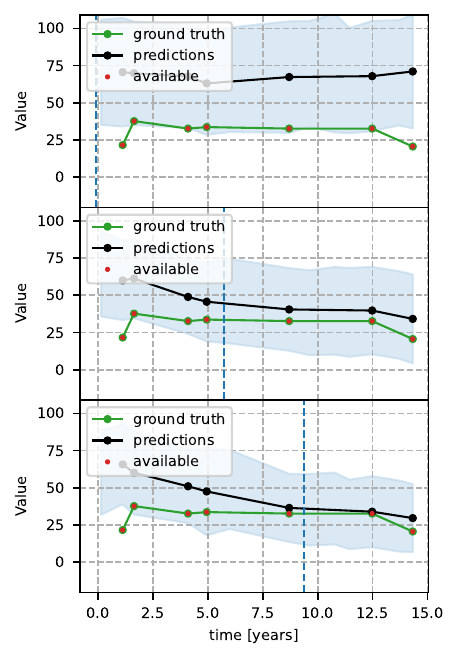}}
\end{figure}
Figure \autoref{fig:stage_pat} shows predicted probabilities of organ stages at a given time point. The intensity of the heatmap reflects the predicted probability.

  
    

\begin{figure}[htbp]
\floatconts{fig:stage_pat}
  {\caption{Probabilities of organ stages for $p_{idx}$. }}
{\includegraphics[width=0.7\linewidth]{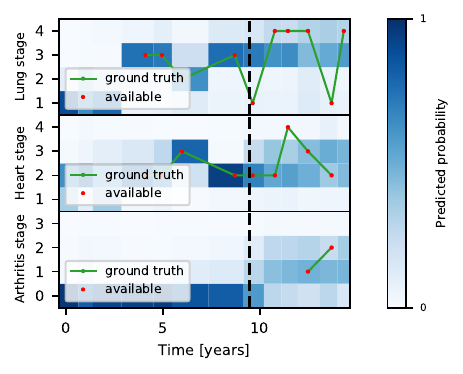}}
\end{figure}
\subsection{Cohort Analysis}
We present here additional cohort-level experiments using our model. 
\subsubsection{Prior z Distributions}
\label{sec:app:prior}

\begin{figure}[t]
\floatconts
  {fig:prior_pred}
  {\caption{Prior predicted $\bs{x}$ trajectories conditioned on time and static variables.}}
  {%
  
    \subfigure[Prior FVC trajectories overlaid with different SSC subsets.]{\label{fig:prior_fvc}%
      \includegraphics[width=0.7\linewidth]{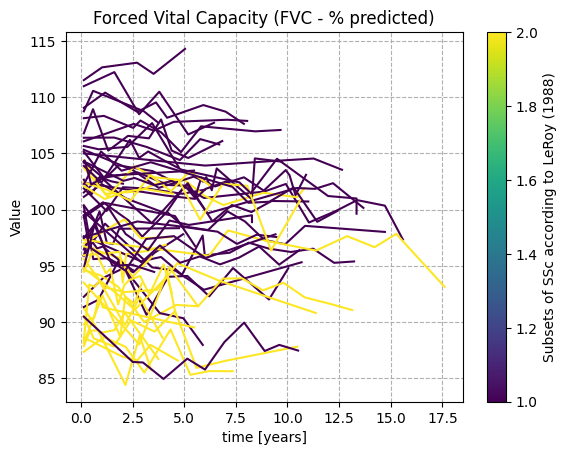}}%
    \qquad
    
    \subfigure[Prior natriuretic peptides trajectories overlaid with date of birth.]{\label{fig:ntb_prior}%
      \includegraphics[width=0.7\linewidth]{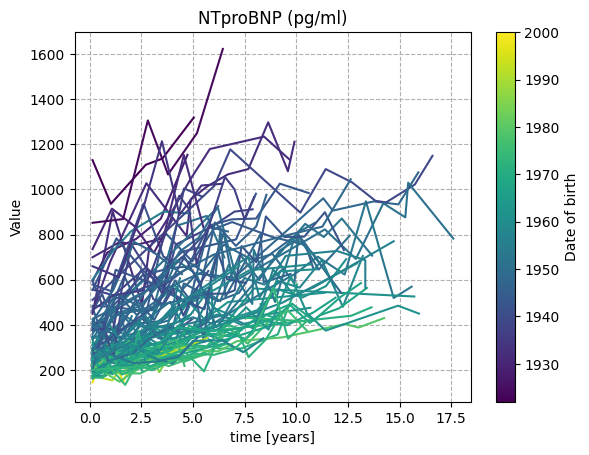}}
  }
\end{figure}

By learning $p(\bs{x},\bs{y} \vert \bs{s}, \bs{\tau})$, we estimate the average prior disease trajectories in the cohort. This allows the comparison of trajectories, conditioned only on the simple subset of variables $\bs{s}$ and $\bs{\tau}$ and thus without facing any confounding in the trajectories, for instance, due to past clinical measurements $\bs{x}$. For example, in  Figure \autoref{fig:prior_fvc} we overlaid the predicted prior trajectories of Forced Vital Capacity (FVC)\footnote{FVC is the amount of air that can be exhaled from the lungs. Low levels indicate lung malfunction.} for a subset of patients in $\mathcal{P}_{test}$ with a static variable corresponding to the SSc subtype.  Overall, the FVC values are predicted to remain quite stable over time, but with different average values depending on the SSc subtype. In  Figure \autoref{fig:ntb_prior}, the prior predicted N-terminal pro b-type natriuretic peptide (NTproBNP)\footnote{They are substances produced by the heart. High levels indicate potential heart failure.} trajectories overlaid with age, show that the model predicts an overall increase in NTproBNP over time, and steeper for older patients.
\subsubsection{Latent Space and Medical Concepts}
\begin{figure*}[htbp]
    \floatconts
    {fig:tsnes_stages}
    {\caption{Predicted organ stages in the latent space. The red line highlights the trajectory of $p_{idx}$.}}
    {\includegraphics[width=0.9\linewidth]{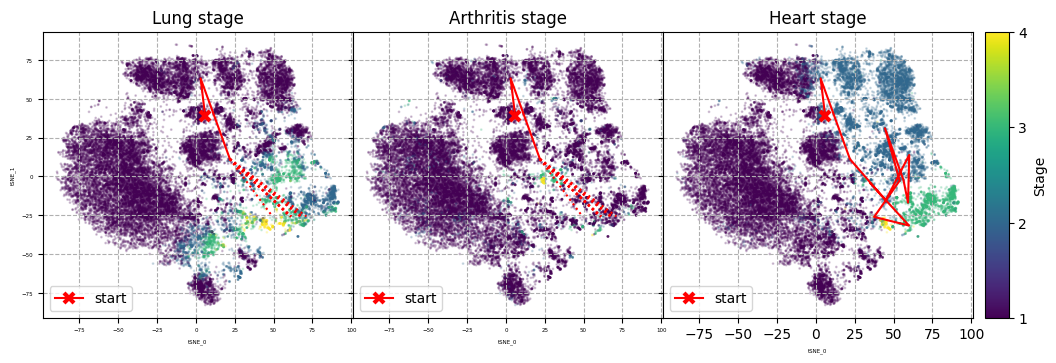}}
\end{figure*}
\label{sec:app:latent}
\paragraph{$\bf{t-}$SNEs:}{
\label{sec:app:tsne}
The $t$-SNE \citep{van2008visualizing} graphs were obtained by computing the two-dimensional $t$-SNE projection of the latent variables  $\bs{z}_{1:T} \mid (\bs{x}_{1:T}, \bs{c})$ (i.e. only using reconstructed $\bs{z}$) of a subset of $\mathcal{P}_{train}$ and then transforming and plotting the projected latent variables (reconstructed or forecasted) from patients in $\mathcal{P}_{test}$ \citep{polivcar2019opentsne}}. \\

In \autoref{fig:tsnes}, we showed the trajectory of $p_{idx}$ overlaid with the predicted organ involvement probabilities. In \autoref{fig:tsnes_stages}, we additionally show the trajectory overlaid with the organ stages, showing for instance in the first panel that the model predicts an increase in the lung stage and in the last panel that $p_{idx}$ undergoes many different heart stages throughout the disease course.

\subsubsection{Clustering of Patient Trajectories and Trajectory Similarity}
We discuss additional results obtained through clustering and similarity analysis of latent trajectories (\autoref{sec:clustering}). 
In \autoref{fig:clust_traj_proba}, we show the different predicted probabilities of the medical concepts $\bs{y}$ for the mean trajectories within the three found clusters. This reveals which medical concepts are most differentiated by the clustering algorithm. For instance, cluster one exhibits low probabilities of organ involvement, while cluster two shows increasing probabilities of heart involvement and low probabilities of lung involvement. In contrast, cluster three shows increasing probabilities for both heart and lung involvement.

\begin{figure*}[htbp]
\centering
\floatconts{fig:tsne_traj_p_simil}
  {\caption{Trajectory of $p_{idx}$ and its $3$ nearest neighbors in the latent space. }}
{\includegraphics[width=0.7\linewidth]{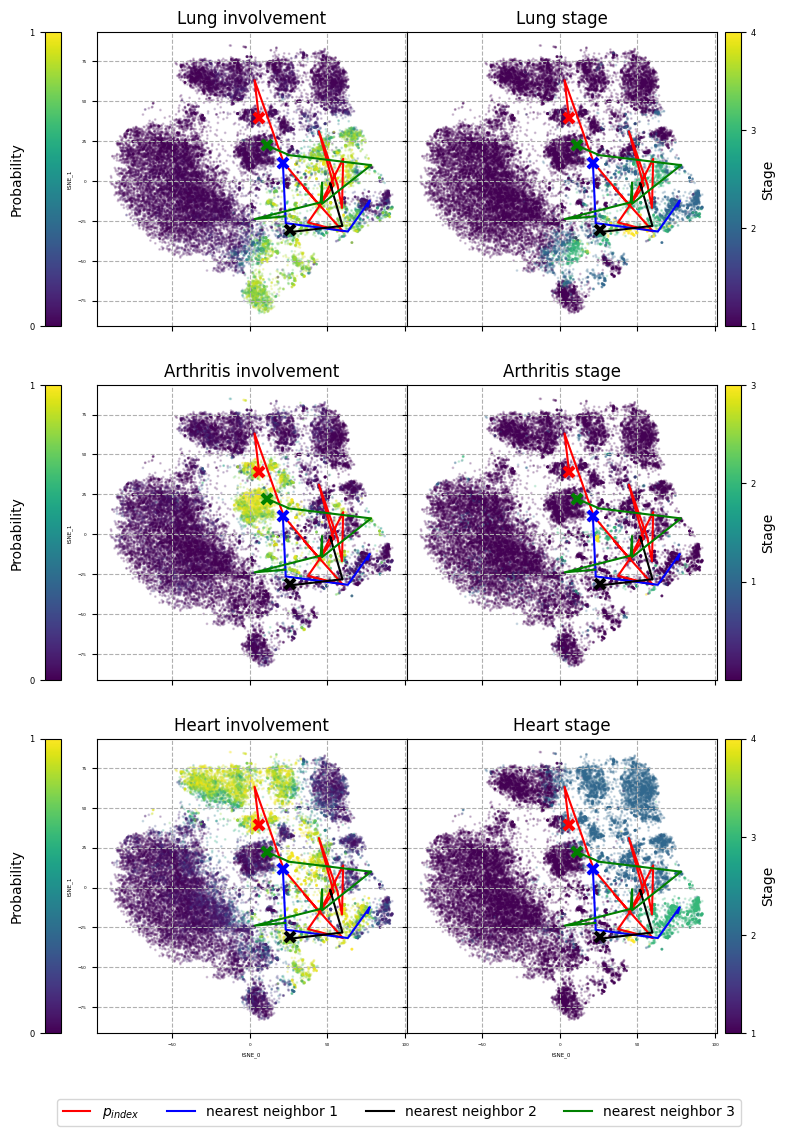}}
\end{figure*}
Additionally, we apply a \emph{k-nn} algorithm with the \emph{dtw} distance in the latent space to find patients with similar trajectories to $p_{idx}$. \autoref{fig:tsne_traj_p_simil} shows the trajectory of $p_{idx}$ and its three nearest neighbors in the latent space. We can see that the nearest neighbors also have an evolving disease, going through various organ involvements and stages. Similarly, in \autoref{fig:traj_p_simil}, the medical concept trajectories of $p_{idx}$ and its nearest neighbors reveal consistent patterns. 
\label{sec:app:clust}

\begin{figure}[htbp]
\centering
\floatconts{fig:clust_traj_proba}
  {\caption{Medical concept trajectories for cluster means. }}
{\includegraphics[width=0.9\linewidth]{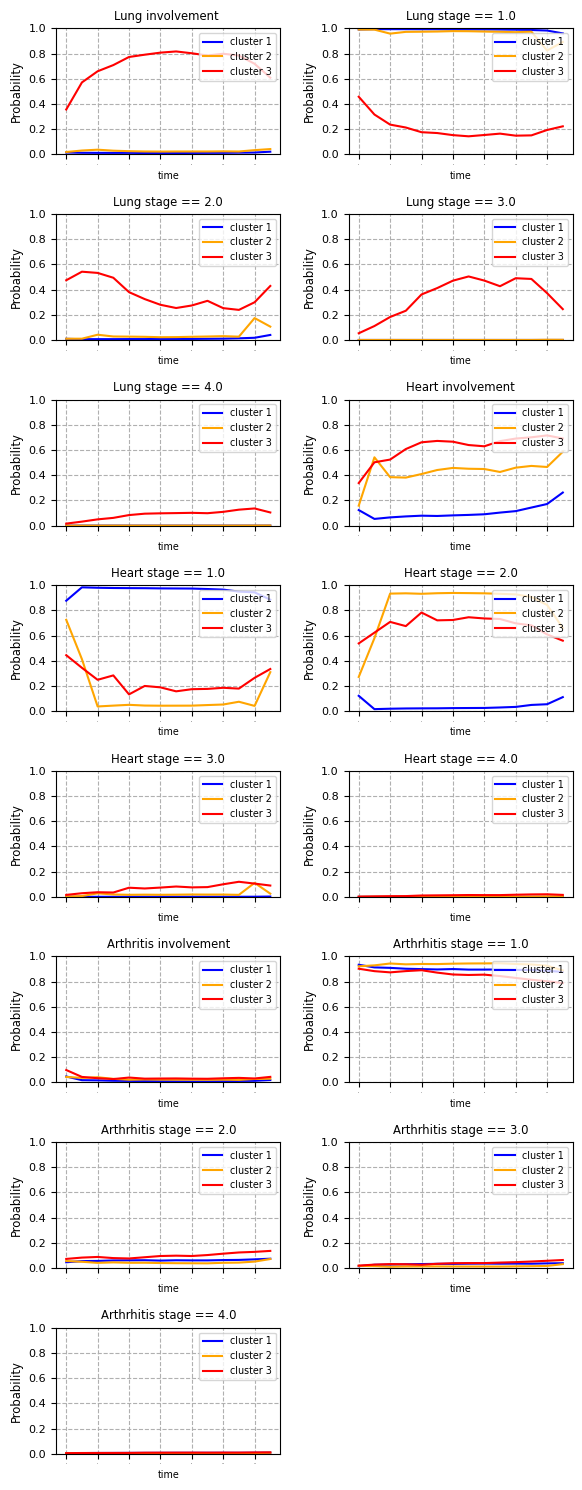}}
\end{figure}

\begin{figure}[htbp]
\centering
\floatconts{fig:traj_p_simil}
  {\caption{Medical concept trajectories for $p_{idx}$ and its $3$ nearest neighbors. }}
{\includegraphics[width=0.9\linewidth]{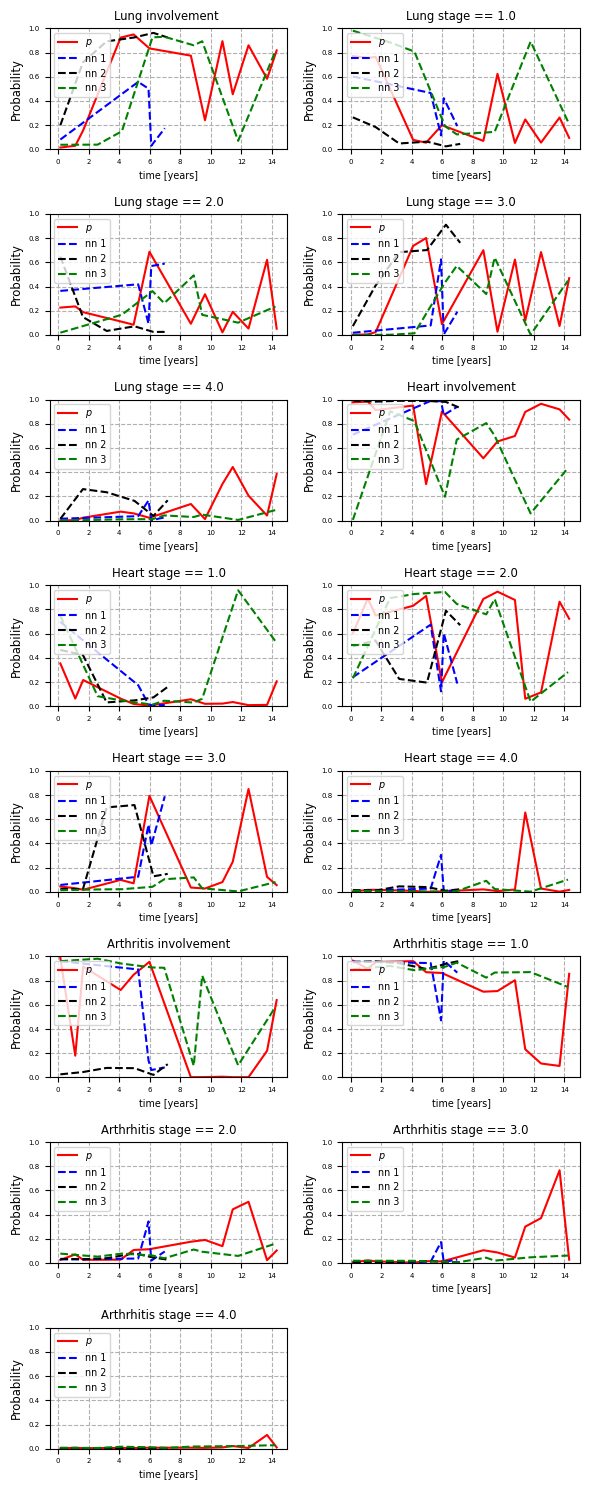}}
\end{figure}

\clearpage
\section{Deep Generative Temporal Models}
\label{sec:deep_gen}
In this section, we provide an extended background about different aspects of deep generative temporal models.

\subsection{Probabilistic Generative Models}
Probabilistic generative models differ from deterministic discriminative models in two key aspects. 
First, 
probabilistic  models provide richer information by learning not only the mean $\mathbb{E}_{p(\bs{y})}[\bs{y}]$ but also the complete distribution $p(\bs{y})$, which can be used for instance for uncertainty quantification. 
Second, generative models learn the entire distribution $p(\bs{y}, \bs{x})=p(\bs{y}|\bs{x})p(\bs{x})$, capturing the joint distribution over all involved variables $\bs{y}$ and $\bs{x}$. In contrast, discriminative models only learn the conditional distribution $p(\bs{y}|\bs{x})$ by separating target and output variables, without gaining knowledge about $\bs{x}$. Generative models allow to generate new samples from the learned distribution, offering a more holistic approach to modeling data. There are several other deep generative approaches besides latent variable models, for which we refer to \cite{Tomczak2022DeepModeling}.
%

\subsection{Conditional Generative Models}
Generative models can be enhanced by incorporating additional context variables $\bs{c}$, leading to the conditional generative model 
$p( \bs{y}, \bs{x} \vert \bs{c}) 
=
p( \bs{y} \vert \bs{x}, \bs{c}) p( \bs{x} \vert \bs{c}),
$
which could for instance model the interplay between the variables $\bs{y}$ and $\bs{x}$ given the personalized context $\bs{c}$. This distribution can be useful to generate novel realistic samples of $\bs{y}$-$\bs{x}$ pairs, conditioned on the individual past context or a specific condition.

\subsection{Latent Variable Models}
Probabilistic generative models can be extended by some latent variables $\bs{z}$, which reflects the assumption that not all variables are observed. This allows to model a generative process  involving the observed variables  $\bs{y}$ and  $\bs{x}$, as well as   some
latent variables $\bs{z}$ by modeling the joint distribution
$p( \bs{y}, \bs{x}, \bs{z} )$, where the latent variables are learned during inference and marginalized out for prediction or other queries. These probabilistic representations can be used for identifying new patterns or concepts, not explicitly present in the data. Furthermore, they also allow modeling much more complex distributions beyond standard parametric distributions.

\subsection{Temporal Generative Models}
In the context of generative models, the variables can be represented as vectors or matrices, with the option to explicitly index them by time, denoted as $\bs{x}_t$. This temporal indexing enables the modeling of dynamical systems or time series $\bs{x}_{1:T} = [\bs{x}_1,\ldots, \bs{x}_t,\ldots,\bs{x}_T]$. Various approaches exist for modeling temporal data, such as discrete-time and continuous-time models, as well as deterministic and stochastic dynamic process models. The key distinction lies in how the variables relate to each other over time. For discrete-time models, the variables are updated at specific time intervals, making them suitable for systems that evolve in distinct steps. On the other hand, continuous-time models allow for a more seamless representation of systems that change continuously over time or are irregularly measured, providing a more accurate description of certain processes. Deterministic dynamic process models assume that the relationship between all time steps is completely predictable and certain. For instance, given the initial $\bs{x}_{1}$ state, all future steps are completely determined by a deterministic process. In contrast, probabilistic dynamic process models introduce an element of uncertainty, thus assuming that the evolution of the system from $\bs{x}_{t-1}$ to $\bs{x}_t$ is subject to probabilistic influences.

\subsection{Deep Probabilistic Generative Models}
Deep generative models constitute a very powerful class of probabilistic generative models, where deep neural networks (powerful function approximators) are used to parameterize distributions. Consider a parametric conditional distribution $p(\bs{y}|\bs{x})$ with parameters $\mu_j$ (for instance the mean and standard deviation in a Gaussian). A deep neural network can be used to learn these parameters $\mu_j(\bs{x}, \theta) = g_{\theta}(\bs{x})$ with a deep relation involving some learnable parameters $\theta$. This allows specifying and learning complex conditional parameterized distributions $p_{\theta}(\bs{y}|\bs{x})$. Importantly, although the building blocks are (deep) parametric distributions, the resulting distributions after inference can be arbitrarily complex and diverge from all parametric distributions.

\end{document}